\documentclass[runningheads]{llncs}
\usepackage[utf8]{inputenc}

\usepackage{graphicx}
\usepackage{amsfonts}
\usepackage{hyperref}
\usepackage{url}
\usepackage{booktabs}
\usepackage{nicefrac}
\usepackage{microtype}
\usepackage{comment}
\usepackage{lipsum}
\graphicspath{ {./} }
\usepackage[ruled,vlined,linesnumbered]{algorithm2e}
\usepackage{mathtools}
\usepackage{xcolor}
\usepackage{array}
\usepackage{cleveref}
\usepackage{enumitem}
\usepackage{wrapfig}
\usepackage{booktabs, tabularx}
\usepackage{mathtools,trimclip,stackengine,scalerel}
\usepackage{eucal}
\usepackage{caption}
\graphicspath{ {./images/} }
\usepackage{ifthen}
\usepackage{amsmath}
\usepackage{amssymb}
\usepackage{comment}
\usepackage{thmtools, thm-restate}
\usepackage{hyperref}

\usepackage{environ}

\NewEnviron{sourcefigure}[1][htbp]{
  {\let\caption\relax\let\ref\relax
   \renewcommand{\label}[1]{
    \gdef\sfname{sf:##1}}
    \setbox1=\hbox{\BODY}}
    \global\expandafter\let\csname\sfname\endcsname\BODY
  \begin{figure}[#1]
    \BODY
  \end{figure}
}
\newcommand{\reusefigure}[2][htbp]{
  {\addtocounter{figure}{-1}
   \renewcommand{\theHfigure}{dupe-fig}
   \renewcommand{\thefigure}{\ref{#2}}
   \renewcommand{\addcontentsline}[3]{}
   \renewcommand{\label}[1]{}
   \begin{figure}[#1] \csname sf:#2\endcsname \end{figure}}
}

\SetKwRepeat{Do}{do}{while}
\declaretheorem{Observation}
\newsavebox\tmpbox
\newcommand\hvec[1]{\ThisStyle{%
  \setbox0=\hbox{$\SavedStyle#1$}
  \setbox2=\hbox{$%
    \clipbox{0pt{} \dimexpr\ht0+1.68\LMpt{} -.2\LMpt{} 0pt}{%
      $\SavedStyle\mathaccent"017E{\phantom{\SavedStyle #1}}$}\kern-.2\LMpt$}
    \ensurestackMath{\stackengine{1.3\LMpt}{\SavedStyle#1}{\copy2}{O}{c}{F}{F}{S}}
}}
\SetKwInOut{Parameter}{Parameters}
\SetKwInOut{Parameter}{Static Parameters}
\title{Boosting Robustness Verification of Semantic Feature Neighborhoods}

\author{Anan Kabaha\inst{1} \and
Dana Drachsler-Cohen\inst{1}}
\authorrunning{A. Kabaha and D. Drachsler-Cohen}
\institute{
Technion, Haifa, Israel
\email{\{anan.kabaha@campus,ddana@ee\}.technion.ac.il}}

\newcommand{\tool}{VeeP\xspace}
\begin{document}
\maketitle
\begin{abstract}
Deep neural networks have been shown to be vulnerable to adversarial attacks that perturb inputs based on semantic features.
Existing robustness analyzers can
reason about semantic feature neighborhoods to increase the networks' reliability. However, despite the significant progress in these techniques, they still struggle to scale to deep networks and large neighborhoods. In this work, we introduce \tool, an active learning approach that splits the verification process into a series of smaller verification steps, each is submitted to an existing robustness analyzer. The key idea is to build on prior steps to predict the next optimal step. The optimal step is predicted by estimating the robustness analyzer's \emph{velocity} and \emph{sensitivity} via parametric regression. 
We evaluate \tool on MNIST, Fashion-MNIST, CIFAR-10 and ImageNet and show that it can analyze neighborhoods of various features:  brightness, contrast, hue, saturation, and lightness. We show that, on average, given a 90 minute timeout,
 \tool verifies 96\% of the maximally certifiable neighborhoods within 29 minutes, while existing splitting approaches verify, on average, 73\% of the maximally certifiable neighborhoods within 58 minutes.
\end{abstract}

\section{Introduction}
\label{sec:introduction}

\begin{figure}[t]
   \centering
  \includegraphics[width=1.0\linewidth, trim=0 360 80 0, clip,page=1]{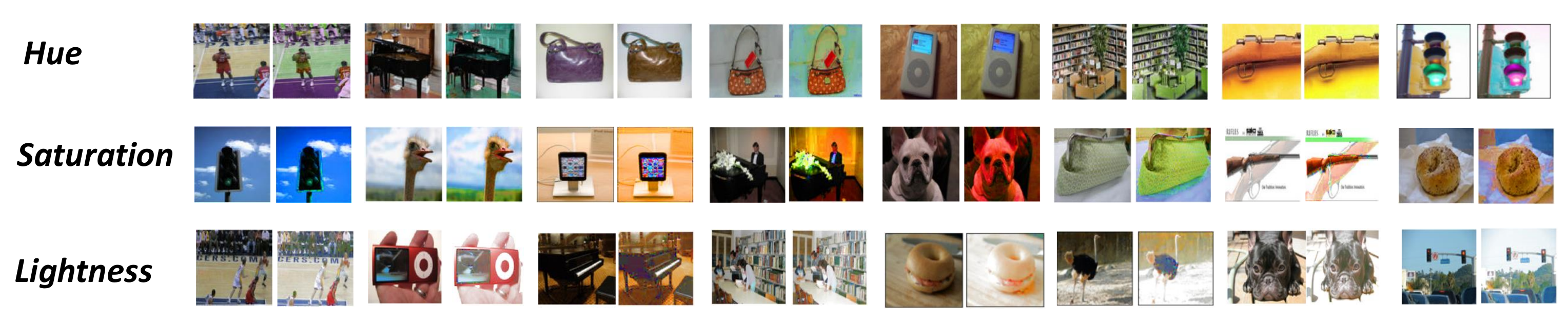}
    \caption{Examples of ImageNet images and maximally perturbed images in the neighborhoods that \tool verified robust, for an AlexNet model.}
    \label{fig:fig_intro}
\end{figure}

The reliability of deep neural networks (DNNs) has been undermined by adversarial examples: perturbations to inputs that deceive the network.
Many adversarial attacks perturb an input image by perturbing each pixel independently by  up to a small constant $\epsilon$~\cite{ref7,ref14,ref55,ref59,ref67}.
 To understand the local robustness of a DNN in $\epsilon$-balls around given images, many analysis techniques have been proposed~\cite{ref120,ElboherGK20,ref18,ref100,ref72,ref73,ref74,ref75,ref35,ref4,ref108}.
In parallel, semantic adversarial attacks have been introduced, such as HSV transformations~\cite{ref77} and colorization and texture attacks~\cite{ref91}. \Cref{fig:fig_intro} illustrates some of these transformations.
Unlike $\epsilon$-ball adversarial attacks which are not visible, feature adversarial attacks can be visible, because the assumption is that humans and networks should not misclassify an image due to perturbations of semantic features.
Reasoning about networks' robustness to semantic feature perturbations introduces new challenges to robustness analyzers. The main challenge is that unlike $\epsilon$-ball attacks, where pixels can be perturbed independently, feature attacks impose dependencies on the pixels. Abstracting a feature neighborhood to its smallest bounding $\epsilon$-ball will lead to too many false alarms. Thus, existing robustness analyzers designed for $\epsilon$-ball neighborhoods perform very poorly on feature neighborhoods.

This gave rise to several works on analyzing the robustness of
 feature neighborhoods~\cite{ref76,ref34,ref35}.
These works rely on existing $\epsilon$-ball robustness analyzers and employ two main techniques to reduce the loss of precision.
First, they
encode the pixels' dependencies imposed by the features by adding layers to the network~\cite{ref76} or by computing a tight linear abstraction of the feature neighborhood~\cite{ref34}.
Second, they split the input range into smaller parts, each is verified independently, e.g., using uniform splitting~\cite{ref76,ref34,ref35}.
 Despite of these techniques, for deep networks and large neighborhoods, existing works either lose too much precision and fail to verify or split the neighborhoods into too many parts. In the latter case, approaches must choose between a very long execution time (several hours for deep networks and a single neighborhood) or forcing the analysis to terminate within a certain timeout, leading to certification of neighborhoods that are significantly smaller than the maximal certifiable neighborhoods. These inherent limitations diminish the ability to understand how vulnerable a network is to feature attacks.

\paragraph{Our work: splitting of feature neighborhoods via active learning}
We address the following problem: given a set of features, each with a target perturbation diameter, find a maximally robust neighborhood defined by these features. We
propose a dynamic close-to-optimal input splitting to boost the robustness certification of feature neighborhoods. Unlike previous splitting techniques, which perform uniform splitting~\cite{ref76,ref34} or branch-and-bound~\cite{ref81,ref100,ref102,ref119,ref120,ref121,ref122}, our splitting relies on active learning: the success or failure of previous splits determines the size of future splits.
The key idea is to phrase the verification task as a process, where each step
picks an unproven part of the neighborhood and submits it to a robustness analyzer.
The analyzer either succeeds in proving robustness or fails.
Our goal is to compute the optimal split.
An optimal split is one where the number of failed steps is minimal, the size of each proven part is maximal, and the execution time is minimal.
Predicting an optimal split requires estimating the exact robustness boundary of the neighborhood, which is challenging.

\paragraph{Splitting by predicting the analyzer's velocity and sensitivity}
We present
\tool (for {\textbf ve}rification {\textbf p}redictor), a learning algorithm, treating the robustness analyzer as the oracle, which dynamically defines the splitting.
\tool defines the next step by predicting the next optimal diameters. To this end, it approximates the analyzer's \emph{sensitivity} and \emph{velocity} for the unproven part. Informally, the sensitivity is a function of the diameters quantifying how certain the robustness analyzer is that the neighborhood is robust. A positive sensitivity means the analyzer determines the neighborhood is robust, while a non-positive sensitivity means the analyzer fails.
The velocity is a function of the diameters quantifying the speed of the robustness analyzer. \tool predicts the diameters of the next step by solving a constrained optimization problem: it looks for the diameters maximizing the velocity such that its sensitivity is positive.
\tool relies on parametric regression to
approximate the velocity and sensitivity functions of the current step.
It terminates either when it succeeds verifying robustness for the given target diameters or when it fails to prove robustness for too small parts. It is thus a sound and precise verifier, up to a tunable precision level.

We implemented \tool in a system, which relies on GPUPoly~\cite{ref72} as the robustness analyzer (the oracle).
We evaluate \tool on different kinds of architectures, including ResNet models for CIFAR-10 and AlexNet models for ImageNet.
~Our experiments focus on several semantic features: brightness, contrast, and HSL (hue, saturation, lightness).
Results show that, when given a 90 minute timeout, \tool almost perfectly closes the gap between the maximal certified feature neighborhoods and the minimal feature adversarial examples:
the verified diameters that \tool computes are, on average, at least $96\%$ of the maximal certifiable diameter. On average, \tool completes in 29 minutes.
We compare to branch-and-bound, which computes 74\% of the maximal diameters in 54 minutes, and to uniform splitting, which computes 73\% of the maximal diameters in 62 minutes.
We study the acceleration rate of \tool over branch-and-bound and uniform splitting by running an experiment without a timeout. Results show that \tool reduces the execution time of branch-and-bound by 4.4x and of uniform splitting by 10.2x.
We also compare to the theoretical optimal greedy baseline that ``knows'' the optimal diameter of every step. We show that \tool's time overhead is only 1.2x more than this theoretical optimal baseline.
\Cref{fig:fig_intro} illustrates how large the neighborhoods that \tool verifies.
It shows pairs of original ImageNet images and the maximally perturbed image in the neighborhood that \tool verified robust, for an AlexNet model. In these examples, every neighborhood is defined by a different feature (hue, saturation, and lightness), and the target diameter submitted to \tool is determined by computing a minimal adversarial feature example along the corresponding feature.

To conclude, our main contributions are:
\begin{itemize}[nosep,nolistsep]
\item A learning algorithm, called \tool, to verify robustness of feature neighborhoods. \tool
computes an optimal split of the neighborhood, each part is verified by a robustness analyzer. To predict the next split, \tool approximates the analyzer's velocity and sensitivity
using parametric regression.
\item An evaluation of \tool on MNIST, Fashion MNIST, CIFAR-10 and ImageNet over fully-connected, convolutional, ResNet, and AlexNet models. Our evaluation focuses on neighborhoods defined using brightness, contrast, and HSL.
    Results show that \tool provides a significant acceleration over branch-and-bound and uniform splitting.
\end{itemize}

\section{Preliminaries}
\label{sec:preliminary}
In this section, we provide the background on neural network classifiers, verification of feature neighborhoods, and existing splitting approaches.

\paragraph{Neural network classifiers} Given an input domain $\mathbb{R}^d$ and a set of classes $C=\{1,\dots,c\}$, a classifier is a function mapping inputs to a score vector over the possible classes $D:\mathbb{R}^d\to \mathbb{R}^c$.
A fully-connected network consists of $L$ layers. The first layer takes as input a vector from $\mathbb{R}^d$, denoted ${i}$, and it passes the input as is to the next layer.
The last layer outputs a vector, denoted $o^D({i})$, consisting of a score for each class in $C$. The classification of the network for input ${i}$ is the class with the highest score, $c'= \text{argmax}({o}^D({i}))$.
When it is clear from the context, we omit the superscript $D$.
 The layers are functions, denoted $h_1, h_2, \dots, h_{L}$, each takes as input the output of the preceding layer. The network's function is the composition of the layers: ${o}({i})=D({i})=h_{L}(h_{L-1}(\cdots(h_1({i}))))$.
 The function of layer $m$ is defined by a set of processing units called neurons, denoted $n_{m,1},\ldots,n_{m,k_m}$. Each neuron
 takes as input the outputs of all neurons in the preceding layer and outputs a real number. 
 The output of the layer ${m}$ is the vector $(n_{m,1},\ldots,n_{m,k_m})^T$ consisting of all its neurons' outputs.
  A neuron $n_{m,k}$ has a weight for each input ${w}_{m,k,k'}$ and a single bias $b_{m,k}$.
  Its function is computed by first computing the sum of the bias and the multiplication of every input by its respective weight:
 $\hat{n}_{m,k}=b_{m,k}+\sum_{k'=1}^{k_{m-1}}{w}_{m,k',k}\cdot{n}_{m-1,k'}$. 
   This output is then passed to an activation function
  $\varphi$ to produce the output $n_{m,k}=\varphi(\hat{n}_{m,k})$.
  Activation functions are typically non-linear functions. In this work, we focus on the ReLU activation function, $\text{ReLU}(x)=\max(0,x)$.
  We note that, for simplicity's sake, we explain our approach for fully-connected networks, but it extends to other architectures, e.g., convolutional and residual networks.

  \paragraph{Local robustness}
A safety property for neural networks that has drawn a lot of interest is \emph{local robustness}. Its meaning is that a network does not change its classification for a given input under a given type of perturbation. Formally,
given an input $x$, a neighborhood containing $x$, $I(x)\subseteq \mathbb{R}^d$, and a classifier $D$, we say $D$ is robust in $I(x)$ if $\forall x'\in I(x)$, $\text{argmax}(D(x'))=\text{argmax}(D(x))$.
We focus on feature neighborhoods, consisting of perturbations of an input $x$ along a set of features $f_1,\ldots,f_T$.
The perturbation of an input along a feature $f$ is a function
 $f:\mathbb{R}^d\times \mathbb{R} \to \mathbb{R}^d$, mapping an input $x$ and a diameter ${\delta}$ to the perturbation of $x$ along the feature $f$ by ${\delta}$.
 To abbreviate, we call the perturbation function the feature $f$, similarly to~\cite{ref76}.
 For all features $f$ and inputs $x$, we assume $f(x,0)=x$.
Given a feature $f$, a diameter $\bar{\delta}$, and an input $x$, the feature neighborhood $I_{f,\bar{\delta}}(x)$ is the set of all perturbations of $x$ along $f$ by up to diameter $\bar{\delta}$:
$I_{f,\bar{\delta}}(x)=\{f(x,\delta)\mid 0\leq \delta\leq \bar{\delta}\}$.
We extend this definition to a set of features by considering a diameter for every feature. 
Given a set of features $f_1,\ldots,f_T$, their diameters $\bar{\delta}_1,...,\bar{\delta}_T$, and an input $x$, we define:
$$I_{f_1,\bar{\delta}_1,...,f_T,\bar{\delta}_T}(x)=\{f_T(... f_2(f_1(x,\delta_1),\delta_2)...,\delta_T)\mid 0\leq \delta_1\leq \bar{\delta}_1,...,0\leq\delta_T\leq \bar{\delta}_T\}$$

\section{Verification of Feature Neighborhoods: Motivation}\label{sec:motiv}
\sloppy
There are many verifiers for analyzing robustness of neural networks~\cite{ref120,ElboherGK20,ref18,ref100,ref72,ref73,ref74,ref75,ref35,ref4,ref108}. Most of them analyze box neighborhoods, where each input entry is bounded by an interval $[l,u]$ (for $l,u\in \mathbb{R}$).
In particular, they can technically reason about feature neighborhoods: first, one has to over-approximate a feature neighborhood $I_{f_1,\bar{\delta}_1,...,f_T,\bar{\delta}_t}(x)$ to a bounding box neighborhood, and then pass the box neighborhood to any of these verifiers. However, this approach loses the dependency between the input entries, imposed by the features, and may result in spurious counterexamples.
To capture the dependencies, a recent work proposes to encode features as a layer and add it to the network as the first layer~\cite{ref76}. 
This has been shown to be effective for various features, such as brightness, hue, saturation, and lightness. 
However, for deep networks and large feature neighborhoods, encoding the dependency is not enough to prove robustness: either the analysis time is too long or the analyzer loses too much precision and fails.
Because feature neighborhoods have
low dimensionality (every feature introduces a single dimension),
divide-and-conquer is a natural choice for scaling the analysis~\cite{ref76,ref34,ref35}.

\paragraph{Divide-and-conquer for feature neighborhoods}
Divide-and-conquer is highly effective for scaling the analysis of
feature neighborhoods. 
The key challenge is computing a useful split.
A branch-and-bound approach (BaB) computes the split lazily~\cite{ref81,ref100,ref102,ref119,ref120,ref121,ref122}.
To illustrate, consider a single feature neighborhood $I_{f,\bar{\delta}}(x)$. A BaB approach begins by analyzing $I_{f,\bar{\delta}}(x)$. If the analysis fails, it splits the neighborhood into two neighborhoods, $I_{f,{\delta}}(x)$ and $I_{f,\bar{\delta}-\delta}(f(x,{\delta}))$. Then, it analyzes each neighborhood separately and continues to split neighborhoods upon failures. As a result, it tends to waste a lot of time on analyzing too large neighborhoods until reaching to suitable-sized neighborhoods. A uniform splitting approach determines a number $m$ and splits the neighborhood into $I_{f,\bar{\delta}/m}(x),\ldots,I_{f,\bar{\delta}/m}(f(x,\bar{\delta}\cdot(m-1) /m))$~\cite{ref76,ref34,ref35}. This approach may still fail for some neighborhoods, due to timeouts or loss in precision, or waste too much time on verifying too small neighborhoods.
This raises the question: \emph{can we dynamically determine a split that minimizes the execution time of the verification?}

\section{Problem Definition: Time-Optimal Feature Verification}

In this section, we define the problem of robustness verification of feature neighborhoods minimizing the execution time.
To simplify notation, the definitions assume a single feature, but they easily extend to multiple features.

We view the robustness analysis of feature neighborhoods as a process. Given a feature neighborhood, the verifier executes a series of steps, dynamically constructed, until reaching the maximal diameter for which the network is robust.
Our verification process relies on a box analyzer $\mathcal{A}$, which can determine the robustness of box neighborhoods.
Every verification step determines the next (sub)neighborhood to verify and invokes the analyzer.
The analyzer $\mathcal{A}$ need not be complete and may fail due to overapproximation error.
That is, given a network and a box neighborhood, $\mathcal{A}$ returns \emph{robust}, \emph{non-robust}, or \emph{unknown}.
Since the goal of the feature verifier is to compute a maximal neighborhood, if $\mathcal{A}$ returns \emph{unknown}, it splits the last neighborhood into smaller neighborhoods. To guarantee that the verification process terminates, if $\mathcal{A}$ fails to verify a feature neighborhood with a diameter up to a predetermined threshold $\delta_{\text{MIN}}$, we assume that this neighborhood is not robust.
Because the feature verifier terminates when reaching the maximal diameter,
the challenge is not to improve its precision but rather to keep its execution time minimal. We next provide formal definitions.

\begin{definition}[Verification Step]
Given a box analyzer $\mathcal{A}$, a classifier $D$, and a feature neighborhood defined by $f$, $\bar{\delta}$ and $x$, a verification step is a pair $(\delta_x,\delta)$, such that  $0\leq \delta_x <\bar{\delta}$ and $0< \delta \leq \bar{\delta}$.
The result of a verification step $(\delta_x,\delta)$ is $\mathcal{A}$'s result for $D$ and $I_{f,\delta}(f(x,\delta_x))$, which is \emph{robust}, \emph{not robust} or \emph{unknown}.
\end{definition}

We next define feature verification sequence, consisting of verification steps.
\begin{definition}[Feature Verification Sequence]
Given a box analyzer $\mathcal{A}$,
a precision level $\delta_\text{MIN}$,
a classifier $D$, and a feature neighborhood defined by $f$, $\bar{\delta}$, and $x$, a feature verification sequence is a sequence of verification steps $s_1,\ldots,s_m$ that verify the maximally robust neighborhood up to $\bar{\delta}$, i.e., either:
\begin{itemize}[nosep,nolistsep]
\item there is no step whose result is \emph{not robust} and, for every $\delta_y\in [0,\bar{\delta}]$, there is a step $s=(\delta_{x},\delta)$, where $\delta_{x}\leq \delta_y \leq\delta_{x}+\delta$, for which $\mathcal{A}$ returns \emph{robust}.
  That is, the verification steps cover all inputs in $I_{f,\bar{\delta}}(x)$, or
  \item
  there is no step whose result is \emph{not robust}, except perhaps
  the last step $s_m=(\delta_{m,x},\delta_m)$ whose result is \emph{unknown} or \emph{not robust} and $\delta_m=\delta_\text{MIN}$.
      For every $\delta_y\in [0,\delta_{m,x}]$, there is a step $s=(\delta_{x},\delta)$, where $\delta_{x}\leq \delta_y \leq\delta_{x}+\delta$, for which $\mathcal{A}$ returns \emph{robust}. That is,
      the verification steps cover all inputs in $I_{f,\delta_{m,x}}(x)$ and we assume there is an adversarial example in $I_{f,\delta_\text{MIN}}(f(x,\delta_{m,x}))$.
\end{itemize}
\end{definition}

Finally, we define the problem of time-optimal feature verification.
To this end, we introduce a notation. Given a verification step $s$, we denote by $t(s)$ the execution time of the analyzer $\mathcal{A}$ on the neighborhood defined by step $s$.
We note that we assume that the time to define a verification step $s=(\delta_x,\delta)$ is negligible with respect to $t(s)$.
Given a feature verification sequence $S=(s_1,\ldots,s_m)$, its execution time is the sum of its steps' execution times: $t(S)=\Sigma_{i=1}^m t(s_i)$.
Our goal is to compute a feature verification sequence minimizing the execution time.

\begin{definition}[Time-Optimal Feature Verification]
Given a box analyzer $\mathcal{A}$ and a feature neighborhood defined by $f$, $\bar{\delta}$ and $x$, a time-optimal feature verification sequence $S$ is one that minimizes the execution time: $\text{argmin}_{S} t(S)$.
\end{definition}
This problem is challenging because divide-and-conquer algorithms have the execution time of a verification step only \emph{after} they invoke $\mathcal{A}$ on that step's neighborhood. Thus, constructing a verification sequence is bound to involve suboptimal choices.
However, we show that it is possible to \emph{predict} the execution time of a (new) verification step based on the execution times of the previous steps. We note that although we focus on analysis of deep neural networks, we believe that predicting verification steps based on prior steps is a more general concept which is applicable to analysis of other machine learning models.

\section{Prediction by Proof Velocity and Sensitivity}\label{sec:velsen}

In this section, we present the key concepts on which we build to predict the verification steps: proof velocity and sensitivity. We show that these can be modeled by parametric functions. We then explain how these functions can be used to predict optimal steps by solving a constrained optimization problem.

\paragraph{Proof velocity} To minimize the execution time of the verification process,
we wish to maximize the \emph{proof velocity}. Proof velocity is the ratio of the neighborhood's \emph{certified diameter} and the time to verify it by the box analyzer $\mathcal{A}$.
In the following, we denote the execution time of step $s=(\delta_x,\delta)$ by $t(s)=t_\mathcal{A}(I_{f,\delta}(f(x,\delta_x)))$. The certified diameter of
this step's neighborhood, denoted $\delta^s_\mathcal{A}$, is equal to $\delta$, if $\mathcal{A}$ returns \emph{robust}, and $0$, if $\mathcal{A}$ returns \emph{non-robust} or \emph{unknown}.

\begin{definition}[Proof Velocity]
Given a box analyzer $\mathcal{A}$, a classifier $D$, a feature neighborhood defined by $f$, $\bar{\delta}$, and $x$, and a verification step $s=(\delta_x,\delta)$,
the proof velocity of  $s$ is:
$V_\mathcal{A}(I_{f,\delta}(f(x,\delta_x)))=\frac{\delta^s_\mathcal{A}}{t(I_{f,\delta}(f(x,\delta_x)))}$.
\end{definition}

\begin{figure}[t]
    \centering
  \includegraphics[width=1\linewidth, trim=0 156 0 0, clip,page=2]{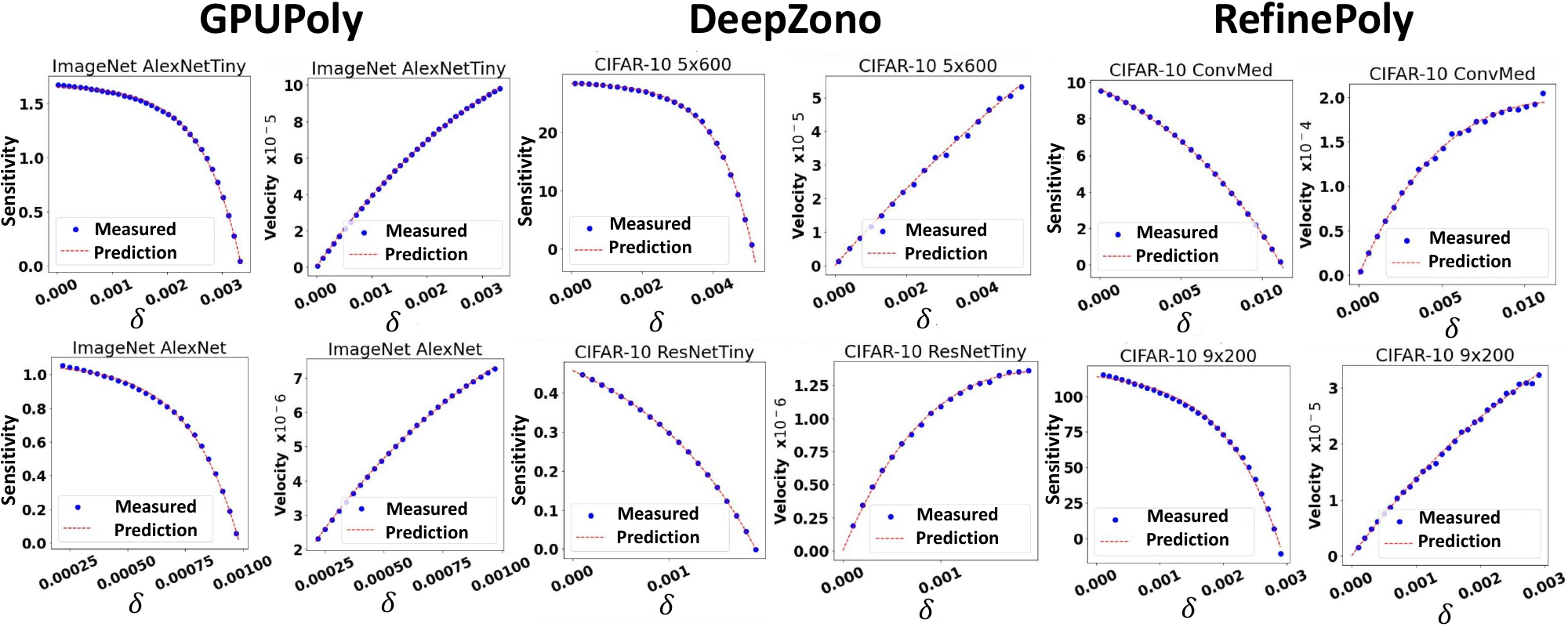}
    \caption{Velocity and sensitivity as functions of the diameter $\delta$, for different models and three box analyzers: GPUPoly~\cite{ref35}, DeepZono~\cite{ref26}, and RefinePoly~\cite{ref89}. Blue dots show the measured values and red curves show our function approximations.}
    \label{fig::model_degradation}
\end{figure}

The velocity is either a positive number, if $\mathcal{A}$ returns \emph{robust}, and $0$ otherwise.
A zero velocity means that the feature verifier has to split this neighborhood
and that we have not gained from this analysis.
Empirically, we observe that if $\mathcal{A}$ relies on linear approximations to analyze the network robustness,
the proof velocity can be modeled as a function of the certified diameter. For small networks or neighborhoods,
the velocity is approximately a linear function of the diameter, because
the analysis time is, in practice, constant.
The larger the network or the neighborhood, the longer the analysis time because the overapproximation error increases, and thus the analyzer $\mathcal{A}$ executes more refinement steps (e.g., back-substitution~\cite{ref35} or solving linear programs~\cite{ref100}). We empirically observe that when the network or the neighborhood are large enough to trigger refinement steps, the execution time is approximately exponentially related to the diameter:
$t(\delta)\propto exp(\beta \cdot \delta)$, for some parameter $\beta$. Consequently, $V(\delta)\propto \delta \cdot exp(-\beta \cdot \delta)$. Note that, for $\beta=0$, the proof velocity is linear in $\delta$. Thus, this function captures both cases of small network/neighborhood and large network/neighborhood.
We illustrate this relation in
\Cref{fig::model_degradation}, showing the measured proof velocity (the blue dots) as a function of the diameter $\delta$, across different models and three box analyzers relying on different linear approximations. The figure also shows the function we use to approximate the proof velocity (the red curve). The figure shows how close the approximation is. We next summarize this observation.

\begin{Observation}
For every verification step $s=(\delta_x,\delta)$, if
  $\delta^s_\mathcal{A}>0$, the velocity can be approximated by: $V(\delta)=\alpha_V \cdot \delta  \cdot exp(-\beta_V  \cdot\delta)$ for $\beta_V\geq0$ and $\alpha_V\in\mathbb{R}$. \label{obs}
\end{Observation}

We can use this observation to predict time-optimal verification steps. To this end, at the beginning of every verification step, we require to (1)~estimate the parameters of the velocity's function and (2)~predict the maximal $\delta^s_\text{MAX}$ for which the analyzer $\mathcal{A}$ returns \emph{robust}.
With these values, we can
 define the next step by computing $\delta\in (0,\delta^s_\text{MAX}]$ maximizing the proof velocity.
 In order to predict the maximal value $\delta^s_\text{MAX}$, we define the concept of \emph{neighborhood sensitivity}.

\paragraph{Neighborhood sensitivity}
The \emph{sensitivity} concept builds on the commonly known concept \emph{network confidence}.
Given a classifier $D$ and an input $x$, the confidence of the classifier in class $j$ is the output ${o}^D_{j}(x)$, i.e., the score that $D$ assigns for $j$ on input $x$. Based on this term, we define the \emph{sensitivity} of $x$ as the difference between the confidence in $j$ and the highest confidence in a class different from $j$:
$$S^D(x,j)= {o}^D_{j}(x) - \text{argmax}_{j'\neq j}({o}^D_{j'}(x))$$
If $S^D(x,j)>0$, then $D$ classifies $x$ as $j$, and
the higher $S^D(x,j)$ the more certain the classifier is in its classification of $x$ as $j$.
 We extend this term to neighborhoods. We define the neighborhood sensitivity
  as the minimal sensitivity of its inputs: $S^D(I,j)=\min\{S^D(x',j)\mid x'\in I\}$.
  From this definition, we get few observations. First, for any $I\subseteq I'$, we have $S^D(I',j)\leq S^D(I,j)$. That is, extending a neighborhood with more inputs may decrease the neighborhood sensitivity in $j$.
  Second, if $S^D(I,j)\leq 0$, then $I$ is not robust to $j$.
  Third, if $\mathcal{A}$ is precise, then for every verification step $s=(\delta_x,\delta)$, we have $\delta^s_\mathcal{A}=\delta$ if and only if
  the sensitivity $S^D(I_{f,\delta}(f(x,\delta_x)),j)$ is positive.
  In practice, we rely on an imprecise analyzer $\mathcal{A}$ and we cannot compute the exact neighborhood sensitivity. However, we can approximate a neighborhood's sensitivity by relying on the analysis of $\mathcal{A}$.
  Since most
  incomplete analyzers compute, for every output neuron $k$, real-valued bounds $[l_{k},u_{k}]$,
  we can approximate the neighborhood sensitivity:
$$S^D_\mathcal{A}(I_{f,\delta}(f(x,\delta_x)),j)=l_j - \max_{j'\neq j} u_{j'}$$
 Thus, to compute the maximal $\delta^s_\text{MAX}$ whose neighborhood can be proven robust by $\mathcal{A}$, we can compute the maximal $\delta^s_\text{MAX}$
 for which $S^D_\mathcal{A}(I_{f,\delta^s_\text{MAX}}(f(x,\delta_x)),j)>0$.
The remaining question is how to approximate the sensitivity function.
Empirically, we observe that if $\mathcal{A}$ relies on linear approximations to analyze the network robustness,
the neighborhood sensitivity has an exponential relation to the diameter.
This is demonstrated in
\Cref{fig::model_degradation}, for different models and linear approximations. The figure shows how close the approximation is (red curves) to the measured sensitivity (the blue dots). We next summarize this observation.

\begin{restatable}[]{Observation}{ftaf}
For every verification step, the neighborhood sensitivity can be approximated by:
$S_\mathcal{A}(\delta)=\alpha_S+\beta_S \cdot exp(\gamma_S \cdot \delta)$, where $\alpha_S,\beta_S,\gamma_S\in \mathbb{R}$.\label{obs2}
\end{restatable}

This exponential relation
can be explained by considering the effect of linear approximations on non-linear computations. At a high-level, the exponential relation is linked to the number of non-linear neurons being approximated. We exemplify this relation in \Cref{sec:appex}.

\paragraph{Time-optimal feature verification via proof velocity and sensitivity}
Given the functions of the velocity and sensitivity, we can state our problem as a constrained optimization.
Given an analyzer $\mathcal{A}$, a feature neighborhood defined by $f$, $\bar{\delta}$ and $x$, and the currently maximal certified diameter $\delta_x$, the $\delta$ of the optimal verification step $s=(\delta_x,\delta)$ is a solution to the following optimization problem:
\begin{equation*}
\begin{multlined}
\phantom{aa}\text{max } V^D(I_{f,\delta}(f(x,\delta_x))) \;
\text{such that } S^D_\mathcal{A}(I_{f,\delta}(f(x,\delta_x)),c_x)> 0
\end{multlined}
\end{equation*}
Here, $c_x$ is the classification of $x$.
Because both functions are convex, the global maximum can be computed as standard.
First, we compute the feasible region of $\delta$ by comparing $S^D(I_{f,\delta}(f(x,\delta_x)),c_x)$ to zero. Second, we compute the derivative of $V^D(I_{f,\delta}(f(x,\delta_x)))$, compare to zero, and compute the optimal $\delta$. If the optimal $\delta$ is not feasible, we take the closest feasible value.
Therefore, if we know the parameters of the velocity and sensitivity functions, we can compute an optimal verification step.
 The challenge is to approximate these parameters, for every step. In the next section, we explain how to predict them from the previous steps.

\section{\tool: A System for Time-Optimal Feature Verification}
In this section, we present our system, called \tool, for computing
time-optimal verification steps.
\tool builds on the ideas presented in \Cref{sec:velsen} and dynamically constructs the verification steps
by solving the constrained optimization problem.
The challenge is predicting the parameters of the
velocity and sensitivity functions.
The key idea is to treat the analyzer as an \emph{oracle},
 whose responses to previous verification steps are used to define the next step.
Conceptually,
\tool builds on active learning, where it acts as the learner for optimal verification steps and
the analyzer acts as the oracle.
Throughout execution, \tool tracks the accumulated verified diameters
of the robust neighborhood. If a verification step succeeds, the robust neighborhood is extended and
the verified diameters increase.
If a step fails, the next predicted diameters will be smaller,
up to a minimal value $\delta_\text{MIN}$.
Thus, although \tool predicts the diameters,
which may be too small or large,
its overall analysis is sound and precise up to $\delta_\text{MIN}$.
It is sound because it employs divide-and-conquer and relies on a sound analyzer.
It is precise because if a step
fails for diameters greater than $\delta_\text{MIN}$, then \tool
attempts again to extend the robust neighborhood by predicting
smaller diameters.
 We begin this section by
explaining how \tool reasons about neighborhoods defined by a single feature
and then extend it to general feature neighborhoods.

\subsection{\tool for Single Feature Neighborhoods}\label{sec:onefeat}
In this section, we describe \tool for analyzing neighborhoods defined by a single feature.
\tool takes as inputs a classifier $D$, a feature $f$, a diameter $\bar{\delta}$, and
an input $x$.
During its execution, it maintains in $\delta_x$ the sum of the certified diameters.
It returns the maximal $\delta_x\leq \bar{\delta}$ for which the neighborhood is robust,
 up to precision $\delta_\text{MIN}$.
\tool operates iteratively, where the main computation of every iteration is
determining a verification step $s_k=(\delta_x,\delta_k)$ to submit to the
 analyzer $\mathcal{A}$.

 \paragraph{Defining a verification step}
The goal of a verification step is to increase the accumulated certified diameter
$\delta_x$ by a diameter $\delta_k$. \tool aims at choosing $\delta_k$ such that (1)~the sensitivity of $I_{f,\delta_k}(f(x,\delta_{x}))$, as determined by the box analyzer $\mathcal{A}$, is positive, and (2)~$I_{f,\delta_k}(f(x,\delta_{x}))$ maximizes the proof velocity.
 \tool leverages Observation~\ref{obs} and~\ref{obs2} and approximates them as
 $S_k(\delta)=\alpha_S+\beta_S \cdot exp(\gamma_S \cdot \delta)$ and
 $V_k(\delta)=\alpha_V \cdot \delta  \cdot exp(-\beta_V  \cdot \delta)$.
 %To determine these functions,
 It solves two parametric regression problems to determine
 $\theta^k_S = (\alpha_S,\beta_S,\gamma_S)$ and $\theta^k_V=(\alpha_V,\beta_V)$.
 This requires to obtain examples:
 $e^1_S=(\delta^1,S(\delta^1)),...,e^M_S=(\delta^M,S(\delta^M))$ and $e^1_V=(\delta^1,V(\delta^1)),...,e^M_V=(\delta^M,V(\delta^M))$.
  The minimal number of examples is
  three for $S_k(\delta)$ and
  two for $V_k(\delta)$.
  Given the examples,
  the parameters are determined by minimizing a loss: %on the provided examples:
  $$\theta_S^k = \underset{\alpha_S,\beta_S,\gamma_S}{\text{argmin}} L(\alpha_S,\beta_S,\gamma_S,e^1_S,\ldots,e^M_S)\hspace{0.9cm}
  \theta_V^k = \underset{\alpha_V,\beta_V}{\text{argmin}}\ L(\alpha_V,\beta_V,e^1_V,\ldots,e^M_V)$$
   For the loss, \tool uses the least squares error.
    Given the parameters, \tool solves the optimization problem (\Cref{sec:velsen})
    to approximate the optimal value of $\delta_k$:
   \begin{equation*}
\begin{multlined}
\phantom{aa}\text{max } {V}_{\theta^k_V}(\delta) \;
\text{such that } S_{\theta^k_S}(\delta)> 0
\end{multlined}
\end{equation*}
The remaining question is how to obtain examples.
A naive approach is to randomly select
$\delta^1,\ldots,\delta^M$ and for each $\delta^i$ run the analyzer $\mathcal{A}$
on $I_{f,\delta^{i}}(f(x,\delta_x))$,
to find the sensitivity and velocity.
However, these $M$ calls to $\mathcal{A}$ are highly time consuming, especially because their only goal is to predict the next diameter to analyze.
Instead, \tool relies on previous steps to estimate examples
by
leveraging two empirical observations. First, the function $V_k(\delta)$ is
similar to previous $V_{k-i}(\delta)$, for
small values of $i$.
Thus, \tool can use as examples  $(\delta_{k-i},V_{k-i}(\delta_{k-i}))$, for small values of $i$.
 Second, the function $S_k(\delta)$ is similar to $S_{k-i}(\delta)$, for
small values of $i$, up to a small alignment term: $S_k(0) - S_{k-i}(0)$.
Thus, \tool can use as examples $(\delta_{k-i},S_{k-i}(\delta_{k-i})
+S_k(0) - S_{k-i}(0))$, for small values of $i$.
Note that computing $S_k(0)$ does not require to run $\mathcal{A}$, because the sensitivity of $I_{f,0}(f(x,\delta_x))$ is exactly the sensitivity of the input $f(x,\delta_{x})$, which can be computed by running it through the classifier $D$.
Based on these observations, \tool obtains examples as follows.
Its first example is $(0,S_k(0))$.
Since the velocity of this step's neighborhood is zero, it is not used to approximate $V_k(\delta)$.
The next $M-1$ examples are defined as described by the previous $M-1$ predicted diameters,
which have already been submitted to $\mathcal{A}$.
Note that the examples are defined from previous steps regardless of whether
their neighborhoods have been proven robust or not.
When \tool begins its computation and has no previous steps, it executes $M-1$ steps whose diameters are some small predetermined values.

 \begin{figure}[t]
    \centering
  \includegraphics[width=1\linewidth, trim=0 290 0 0, clip,page=3]{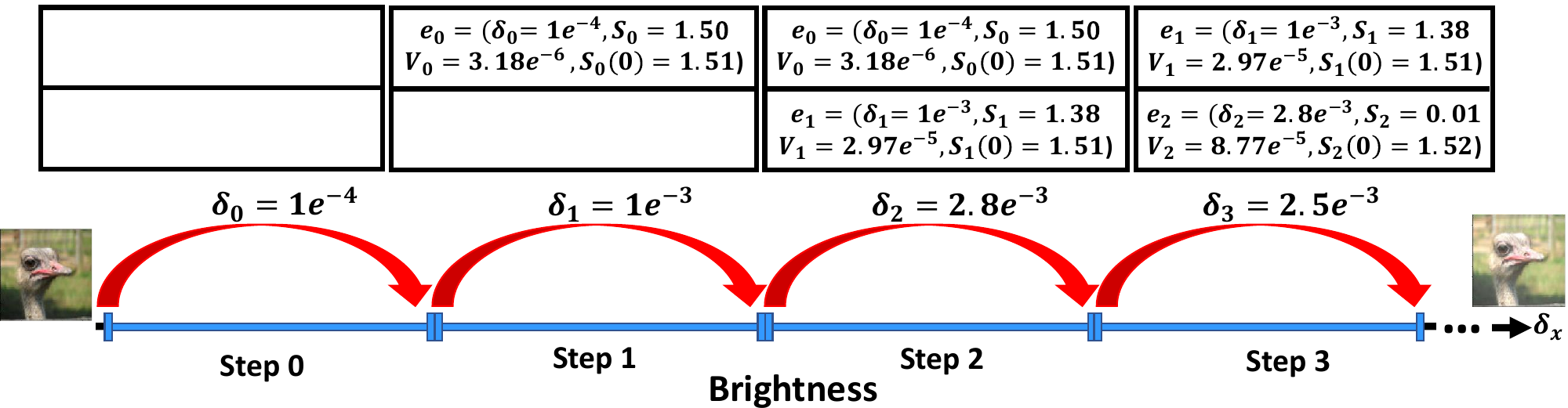}
    \caption{Analysis for
    the brightness feature, an ImageNet image, and AlexNetTiny.}
    \label{fig::example1d11}
\end{figure}

\paragraph{Example} \Cref{fig::example1d11} shows an example of
\tool's analysis for a brightness neighborhood with $\bar{\delta}=0.2$,
an ImageNet image $x$ (the image on the left) and an AlexNetTiny classifier $D$.
We assume $M=3$.
The first two steps rely on predetermined small diameters $\delta_0=10^{-4}$ and $\delta_1=10^{-3}$.
\tool begins by submitting to $\mathcal{A}$ the neighborhood $I_{f,\delta_0}(x)$
 and $\mathcal{A}$ returns
\emph{robust}.
\tool thus updates the accumulated diameter $\delta_x=10^{-4}$
and constructs the example $e_0$. The example consists of
the sensitivity $S_0$ and velocity $V_0$ (computed from $\mathcal{A}$'s analysis), and
the sensitivity $S_0(0)$ at $\delta_x=0$ (computed by
running $x$ through $D$).
The next verification step submits to $\mathcal{A}$ the neighborhood
$I_{f,\delta_1}(f(x,10^{-4}))$ and $\mathcal{A}$ returns \emph{robust}.
\tool thus updates $\delta_x=1.1\cdot10^{-3}$ and
constructs the example $e_1$, consisting of
the sensitivity $S_1$ and velocity $V_1$ (computed from $\mathcal{A}$'s analysis)
and the sensitivity $S_1(0)$
 (computed by running $f(x,10^{-4})$ through $D$).
To predict the next diameter $\delta_2$, \tool relies on $e_0$, $e_1$ and
$S_2(0)$ (computed by running $f(x,1.1\cdot10^{-3})$
through $D$). Its
examples are:
$e_S^0=(0,1.52)$,
$e_S^1=(10^{-4},S_0+S_2(0)-S_0(0))$,
$e_S^2=(10^{-3},S_1+S_2(0)-S_1(0))$, and
$e_V^0=(10^{-4},V_0)$,
$e_V^1=(10^{-3},V_1)$. Given the examples, it minimizes the MSE loss to compute the parameters $\theta_S^2$
and $\theta_V^2$. Afterwards, it solves the constrained optimization function to compute $\delta_2$.
The result is $\delta_2=2.8\cdot 10^{-3}$.
\tool submits to $\mathcal{A}$ the neighborhood $I_{f,\delta_2}(f(x,1.1\cdot 10^{-3}))$
and $\mathcal{A}$ returns \emph{robust}.
\tool updates $\delta_x$ and constructs the example $e_2$, as described before.
\tool
predicts the next diameter $\delta_3$, by repeating this process
using the examples $e_1$ and $e_2$. It continues until reaching the target diameter $\bar{\delta}=0.2$.
The most perturbed image in this neighborhood is shown on the right of \Cref{fig::example1d11}.

\begin{figure}[t]
    \centering
  \includegraphics[width=1\linewidth, trim=0 145 0 0, clip,page=4]{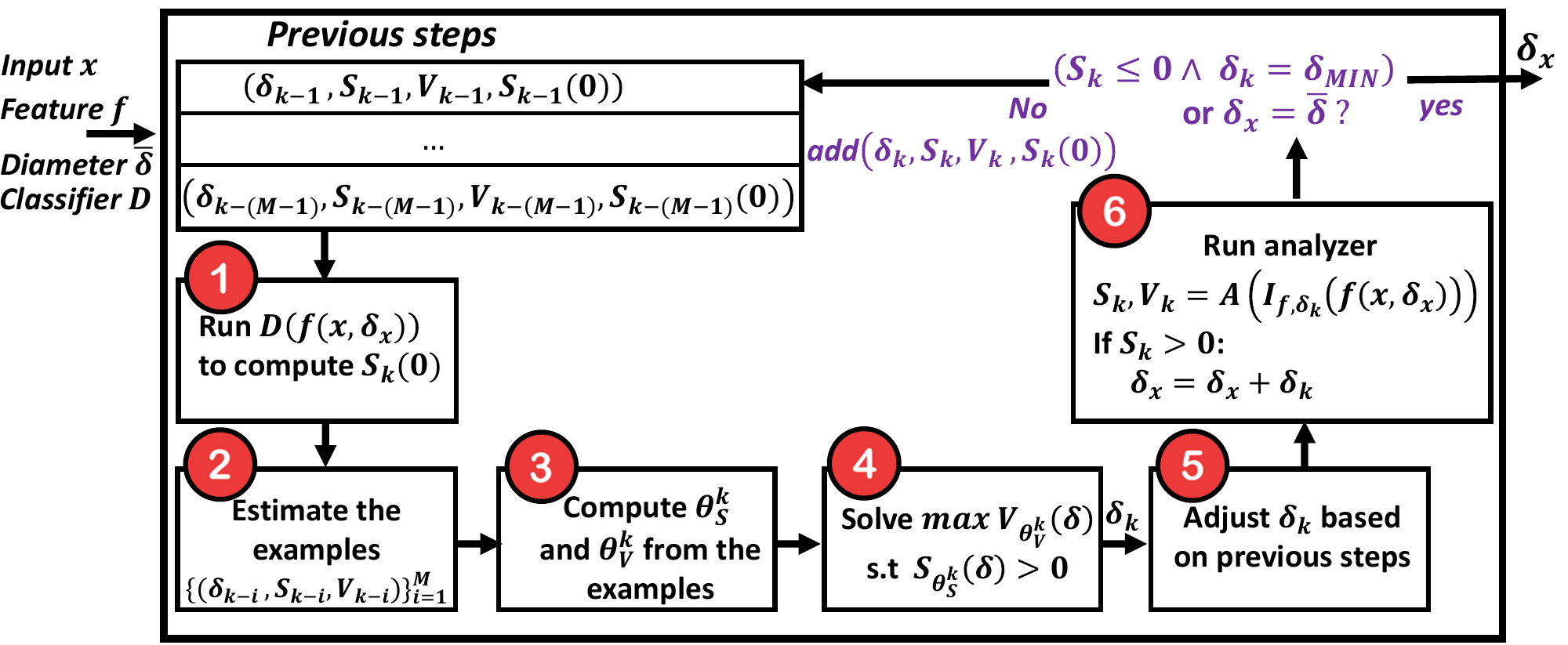}
    \caption{The \tool System.}
    \label{fig:fig_ogis2}
\end{figure}

\paragraph{Overall operation}
The operation of \tool is summarized in \Cref{fig:fig_ogis2} and mostly follows the description above, up to few modifications to guarantee termination.
Initially, \tool sets $\delta_x=0$ and generates the first $M-1$ steps using predetermined diameters.
Every verification step predicts the next diameter based on previous iterations, as described before (steps 1--4 in \Cref{fig:fig_ogis2}). Then, to avoid certification failures and guarantee termination, \tool performs three corrections to the predicted diameter $\delta_k$ (step 5).
First, it checks whether in the last $M$ steps, there has been a step $i$ with a smaller predicted diameter, $\delta_i<\delta_k$, that failed.
If so, \tool
sets $\delta_k$ to the minimal value between the last verified diameter (if exists) and the last failed one
from the last $M$ steps.
Second, it subtracts a small constant from $\delta_k$.
 Third, it guarantees that $\delta_k$ is not below the precision level by
setting $\delta_k=\max(\delta_k, \delta_\text{MIN})$.
These refinements, along with the prediction based on recent examples,
 aim at mitigating predicting too large or too small diameters.
The neighborhood defined by $(\delta_x,\delta_k)$ is submitted to the analyzer $\mathcal{A}$ (step 6),
which returns the real-valued bounds of the output neurons. Accordingly,
\tool computes the sensitivity $S_k$ and velocity $V_k$.
If $S_k>0$, then the neighborhood is robust and thus $\delta_x$ is increased by $\delta_k$.
Afterwards, \tool checks the termination conditions. The first condition is
$S_k\leq 0$ and $\delta_k=\delta_\text{MIN}$, indicating that the neighborhood is maximal.
The second condition is $\delta_x=\bar{\delta}$, indicating that
\tool certified the target diameter. If the conditions are not met, \tool constructs
the example of this step and continues to another iteration.

\paragraph{Correctness analysis}
We next discuss the time overhead of \tool and its correctness.
The first lemma analyzes
the time overhead of \tool.
The overhead is the additional time that \tool requires compared to an oblivious splitting
approach.
The overhead of every step consists of
the call to the classifier $D$ (to compute $S_k(0)$) and
the time to solve the regression problems (to approximate $S_k$ and $V_k$).
The time overhead also includes the $M-1$ initial calls to the analyzer $\mathcal{A}$.

\begin{lemma}
The total overhead is $n\cdot(T_D+T_R)+\Sigma_{i=1}^{M-1} T_{\mathcal{A},i}$,
where $T_D$ is the time to run a single input in the classifier $D$, $T_R$ is the time to solve
 a regression problem from $M$ examples,
$n$ is the number of verification steps and $T_{\mathcal{A},i}$ is the execution time of $\mathcal{A}$
on the $i^\text{th}$ initial step.
\end{lemma}
In practice, $T_D$ and $T_R$ are significantly shorter than the time to run the analyzer $\mathcal{A}$.
Since the value of $M$ is small (we pick $M=3$ or $M=4$), the overhead of the initial queries to the analyzer
is negligible when compared to the total execution time of \tool.
As a result, we observe that the execution time of \tool is very close to the optimal greedy baseline that ``knows'' the optimal diameter of every step.
We continue with a lemma guaranteeing termination and a theorem
guaranteeing soundness and precision (up to $\delta_\text{MIN}$).
Proofs are provided in \Cref{sec:proofs}.

\begin{restatable}[]{lemma}{ftb}
\label{lem2}
Given a classifier $D$, an input $x$, a feature $f$ and a diameter $\bar{\delta}$,
if $\mathcal{A}$ is guaranteed to terminate, then \tool is guaranteed to terminate.
\end{restatable}

\begin{restatable}[]{theorem}{fta}
\label{thm}
Given a classifier $D$, an input $x$, a feature $f$, a diameter $\bar{\delta}$, and a precision level $\delta_\text{MIN}$,
if $\mathcal{A}$ is sound (but may be incomplete), then \tool is:
\begin{itemize}[nosep,nolistsep]
\item sound: if it returns $I_{f,\delta_x}(x)$, then this neighborhood is robust, and
\item precise up to $\delta_\text{MIN}$: if it returns $\delta_x$ smaller than $\bar{\delta}$, then we assume there is $\hat{\delta}\in (\delta_x,\delta_x+\delta_\text{MIN}]$ such that
    $x'=f(x,\hat{\delta})$ is an adversarial example.
\end{itemize}
\end{restatable}

\subsection{\tool for Multi-feature Neighborhoods}
In this section, we present \tool's algorithm to verify neighborhoods defined by multiple
features $f_1,f_2,\ldots,f_T$.
\tool computes a sequence of verification steps that cover the maximal robust
 $T$-dimensional hyper-rectangle neighborhood.
 The sequence is constructed such that \tool computes the maximal diameters
 feature-by-feature.
 To compute the maximal diameter of the $i^\text{th}$ feature, \tool
 computes the maximal robust $i$-dimensional neighborhood of the first $i$ features.
Similarly to \Cref{sec:onefeat},
a verification step is a pair of an offset vector
 $(\delta_1,\ldots,\delta_T)$ (instead of $\delta_x$)
and a diameter $\delta$.
A verification step thus corresponds to a hyper-cube neighborhood
$I_{f_1,\delta,\ldots,f_T,\delta}(x_0)$, where $x_0$ is the perturbation of $x$ as determined by the features
and
offsets ($x_0 = f_T(\ldots(f_2(f_1(x,\delta_1),\delta_2),\ldots),\delta_T)$).
While \tool could predict a different diameter for each feature, this would increase the prediction's complexity by a factor of $T$.
Besides this, the analysis is similar to~\Cref{sec:onefeat} but generalizes
it to high dimension, resulting in few differences.
First, computing the offsets is more subtle than computing $\delta_x$.
Second, the examples used for prediction also leverage the \emph{closest} examples.
Third, computing
the accumulated verified diameters, required for checking the termination conditions,
 involves obtaining
the \emph{vertices} of the certified region. We next explain all these differences,
then exemplify \tool's operation, and finally present the algorithm.

\paragraph{Offsets}
Initially, all offsets are zero. Recall that \tool computes
the maximal diameters feature-by-feature,
and, for every $f_i$, it
 computes the maximal robust $i$-dimensional neighborhood of $f_1,...,f_i$.
After every
verification step, \tool computes the next offsets.
Assume \tool is currently at feature $f_i$.
If a step fails for $\delta>\delta_\text{MIN}$, the offsets of the next step are identical.
If a step fails for $\delta=\delta_\text{MIN}$ or reaches $\bar{\delta}_i$,
\tool computes the initial offsets of $f_{i+1}$, as shortly described.
Otherwise, \tool computes the next offsets based on a feature-by-feature {order} (from $1$ to $i$).
The order, defined in \Cref{sec:appex_offsets}, guarantees that \tool covers the entire $i$-dimensional neighborhood.
We later exemplify it on a running example.
Upon starting a feature $f_j$, \tool computes the initial offsets based on the already certified
neighborhoods. This is obtained by finding the earliest step
forming a vertex on the $j$-dimensional boundary of the certified region, such that the
 vertex's $j^\text{th}$ offset
is within $(0,\bar{\delta}_j)$.
 This leverages the already
 certified neighborhoods: since the
 steps define hyper-cube neighborhoods,
 as a byproduct of their analysis,
 there is also progress in the direction of the succeeding, not yet analyzed, features.
 The complete computation is provided in \Cref{sec:appex_offsets}.

\paragraph{Examples}
The diameter of a verification step is predicted by
  $M+1$ examples: $(0,S_k(0))$, $M-1$ (adapted) recent examples and,
  to increase the prediction accuracy, the
  \emph{closest} example, with respect to the Euclidean distance.
 The $M-1$ recent examples are used only if they (aim to) advance
  the diameter of the same feature as the current step does.
If not all of them advance the same feature,
\tool completes the missing examples with closest examples or initialization examples.

\paragraph{Termination} \tool terminates when it reaches
  all target diameters or all maximal diameters. These conditions generalize the termination conditions
  presented in~\Cref{sec:onefeat}.
  To check the first condition, \tool maintains
an array \texttt{ds} of the certified diameters, which are updated after every verification step.
  The diameters are computed from the vertices bounding the certified region.
  Although the region induced by the maximal diameters is a hyper-rectangle,
  the certified region may form other shapes.
  During the analysis, \tool computes the vertices of the certified region.
To update \texttt{ds}, it selects the maximal bounded
hyper-rectangle, with respect to the Euclidean norm.
  To check the second condition, \tool checks whether it has failed for
  $T$ consecutive iterations for a neighborhood whose diameter is $\delta_\text{MIN}$.
  Correctness follows from the
  the operation of \tool: upon failure of a neighborhood with diameter $\delta_\text{MIN}$,
  it proceeds to the next feature. Thus, $T$ consecutive failures imply that \tool has reached all maximal diameters.

%\begin{figure}[t]
\begin{sourcefigure}[t]
    \centering
  \includegraphics[width=1\linewidth, trim=0 73 175 0, clip,page=5]{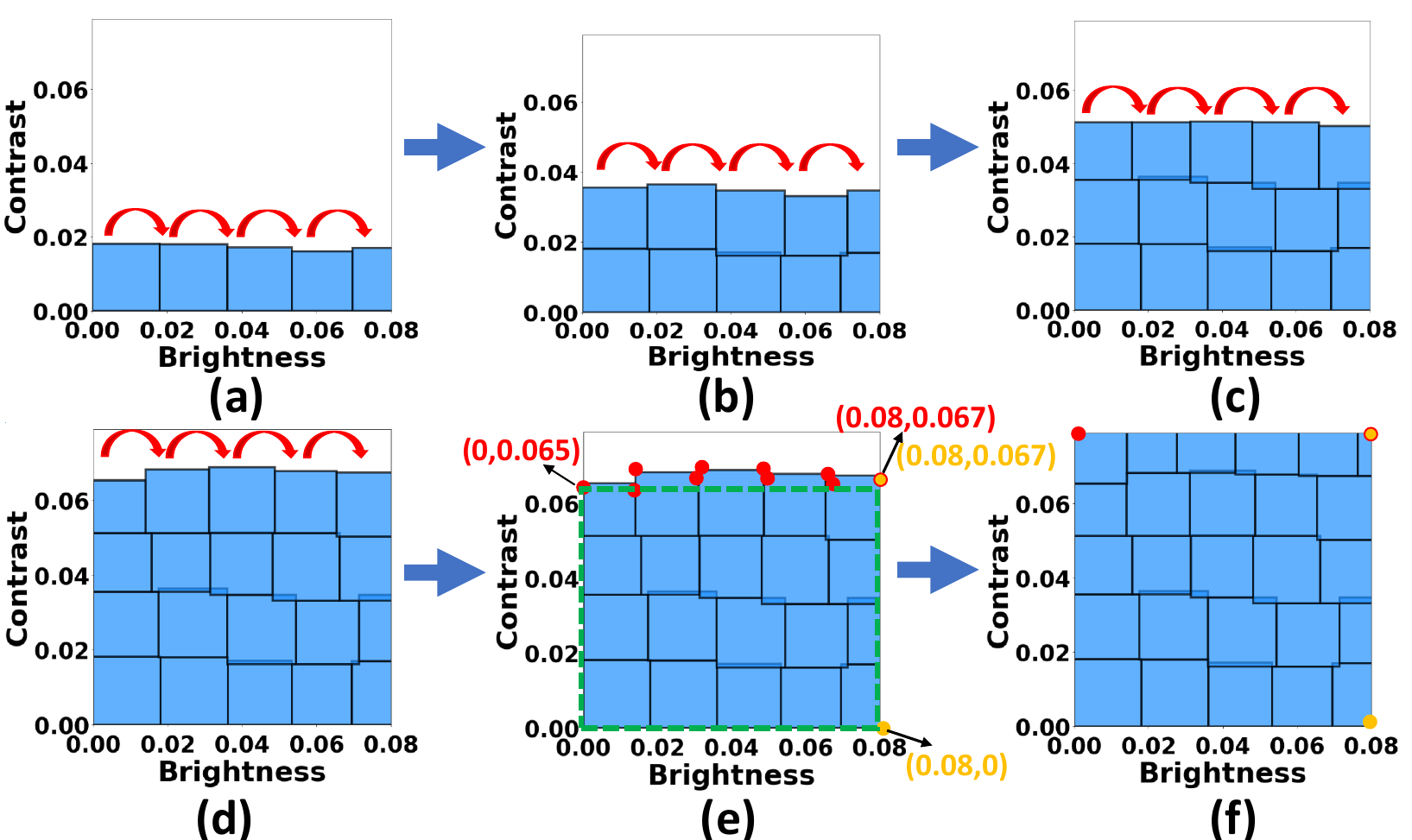}
    \caption{Example of \tool's analysis to certify
    a neighborhood defined by brightness and contrast, for an MNIST image, on a fully-connected network.}
    \label{fig:fig_2d_multi}
%\end{figure}
\end{sourcefigure}

\paragraph{Example} We next exemplify \tool for a
neighborhood defined by brightness and contrast, where $\bar{\delta}_1=\bar{\delta}_2=0.08$
and $M=3$
(\Cref{fig:fig_2d_multi}).
\tool computes the maximal diameters one by one:
first the brightness's diameter and then the contrast's diameter.
\Cref{fig:fig_2d_multi}(a) visualizes the verification steps that compute the maximal
diameter of brightness.
The sequence begins from the offset $(0,0)$ (i.e., $x_0=x$), and the computation
is similar to \Cref{sec:onefeat}.
When \tool reaches $\bar{\delta}_1$, it continues to the contrast feature.
It begins by finding the earliest verification step forming a vertex
on the $2$-dimensional boundary of the certified region, such that the vertex's second offset
is within $(0,\bar{\delta}_2)$. This is the first step and the vertex is $(0,0.018)$
(since this step's diameter is $0.018$).
Thus, the initial offset of contrast is $(0, 0.018)$.
During the analysis of the contrast feature, \tool computes verification
steps feature-by-feature.
Thus, after initializing the offsets, \tool
    advances the brightness's offset,
  until reaching its
  maximal certified diameter (rightmost square, top row, \Cref{fig:fig_2d_multi}(b)).
   Then, by the order \tool follows for the verification steps, it (again) looks for the earliest step
  forming a vertex on the $2$-dimensional boundary of the certified region,
  such that the vertex's second offset is within $(0,\bar{\delta}_2)$.
  This is the leftmost square, top row, \Cref{fig:fig_2d_multi}(b).
  Thus, it sets the next offset (i.e., of the leftmost square, top row,
  \Cref{fig:fig_2d_multi}(c)) to that vertex's offsets.
  The rest of the computation continues similarly
  (\Cref{fig:fig_2d_multi}(c), (d), and (f)).
  We next illustrate the different sets of examples used for the prediction (besides
   $(0,S_k(0))$).
Consider \Cref{fig:fig_2d_multi}(b). The examples used by the middle step at the
 top row are the two leftmost squares at the top row and the middle square at the
 row below.
 The examples used by the leftmost square at the top row are the
 three closest examples -- the three leftmost squares at the bottom row --
 since there are no steps advancing the
 contrast's diameter.
After every verification step,
 \tool constructs for each feature the vertices of the certified neighborhood.
  \Cref{fig:fig_2d_multi}(e) shows the vertices after completing the verification steps of
 \Cref{fig:fig_2d_multi}(d):
 ten red vertices for contrast and two yellow vertices for brightness.
 \Cref{fig:fig_2d_multi}(f) shows the vertices after completing all verification steps.
   Given the vertices, \tool
  computes the accumulated verified diameter of each feature, which is the minimum
  coordinate of its vertices.
 For example, in \Cref{fig:fig_2d_multi}(e), the verified diameter of
 brightness
 is $0.08$, which is the minimum of the first coordinates of $(0.08,0)$ and $(0.08,0.067)$,
 and similarly, the verified diameter of contrast is $0.065$.
  \tool updates the current maximal diameters to these diameters if they form
  a larger hyper-rectangle than the current ones.
Note that if \tool terminates after
reaching all target diameters (e.g., \Cref{fig:fig_2d_multi}(f)),
the certified region is a hyper-rectangle and is thus returned.

\begin{algorithm}[t]
\DontPrintSemicolon
\KwIn{A classifier $D$, input $x$, features $f_1,\ldots,f_T$ and diameters $\bar{\delta}_1,\ldots,\bar{\delta}_T$.}
\KwOut{Diameter array $ds$ s.t. $I_{f_1,ds[1],...,f_T,ds[T]}(x)$ is maximally robust.}
$ds = [0,\ldots,0]$\;
Ex = InitExamples($M$)\;
$\text{offsets} =
[0,\ldots,0]$\;
count\_min = 0\;
\While{$\exists ds[i] < \bar{\delta}_i\wedge count\_min < T
$}{
    $x_0$ = $perturb(x,f_1,\ldots,f_T,\text{offsets})$\;
    $S_0$ = $D(x_0)$ \;
    $\delta$ = predict(Ex, $x_0$, $S_0$)\;
    $t_0 = time()$\;
    $\{l_{o,j},u_{o,j}\}_{j=1}^c$ = $\mathcal{A}(D,I_{f_1,\delta,\ldots,f_T,\delta}(x_0))$ \;
    $t_1 = time()$\;
    $S=l_{c_x}-\max_{j\neq c_x} u_j$\;
    $V = S>0\ ?\ \frac{\delta }{t_1-t_0} : 0$\;
    Ex = Ex $\cup$ $\{(\delta,S,V,S_0,\text{offsets})\}$\;
    $\text{offsets} = \text{compute\_next\_offsets}(Ex,\bar{\delta}_1,\ldots,\bar{\delta}_T)$\;
    $BV_1,\ldots,BV_T = \text{compute\_certified\_neighborhood\_vertices}(Ex)$\;
    $ds\_curr = [0,\ldots,0]$\;
    \For{$i=1$; $i \leq T$; $i++$}{
        $ds\_curr[i] = min\{v_i \mid v\in BV_i\}$ \;
    }
    \lIf{vectorNorm(ds\_curr) $>$ vectorNorm(ds)}{ds = ds\_curr}
    count\_min = ($S \leq 0$ $\wedge$ $\delta$ == $\delta_\text{MIN}$)? count\_min + 1\ :\ 0\;
 }
\Return{$ds$}
  \caption{Multi-feature-\tool($D$, $x$, $f_1$, $\bar{\delta}_1$,\ldots, $f_T$, $\bar{\delta}_T$)}\label{alg:main}
\end{algorithm}

\paragraph{Overall operation}
\Cref{alg:main} summarizes
the operation of \tool.
\tool begins by initializing \texttt{ds}, the maximal
diameters array, the first $M-1$ examples (as described in \Cref{sec:onefeat}), the \texttt{offset} array and a counter \texttt{count\_min}, tracking the number of consecutive failures.
Then, it enters a loop, where each iteration computes a single verification step.
An iteration of the loop begins by determining $x_0$ from the offsets (Line 6).
Then, it progresses as described in \Cref{sec:onefeat} (Lines 7--14):
it computes $x_0$'s sensitivity, predicts $\delta$, submits to $\mathcal{A}$, computes
the velocity and sensitivity, and adds
this verification step as a new example.
After that, it computes the new offsets (Line 15).
Next, the maximal diameters are computed.
To this end,
 \tool constructs, for each feature, the vertices of the certified region (Line 16).
  Computing the vertices is a technical computation determined from the set of examples. We
  omit the exact computation.
  Given the vertices, \tool
  computes the current verified diameters \texttt{ds\_curr}.
 The current verified diameter of feature $i$ is the minimum $i^\text{th}$ coordinate of
 its vertices (Lines 17--19).
 Then, if the Euclidean norm of \texttt{ds\_curr} is greater
than that of \texttt{ds}, it updates \texttt{ds} (Line 20).
Lastly, the counter \texttt{count\_min} is increased, if $\mathcal{A}$ failed, or resets, otherwise (Line 21).
The loop continues as long as \tool has not reached all target diameters and
has not failed during the
last $T$ iterations (Line 5).

 \paragraph{Correctness} We next present the correctness guarantees of \Cref{alg:main}.
 Proofs are provided in \Cref{sec:proofs}.

\begin{restatable}[]{lemma}{ftc}
\label{lem3}
Given a classifier $D$, an input $x$, features $f_1,\ldots,f_T$ and diameters $\bar{\delta}_1,\ldots,\bar{\delta}_T$,
if $\mathcal{A}$ is guaranteed to terminate, then \tool is guaranteed to terminate.
\end{restatable}

Lastly, we show that \tool is sound and precise, up to precision of $\delta_\text{MIN}$ for each feature's maximal diameter.
\begin{restatable}[]{theorem}{ftd}
\label{thm2}
Given a classifier $D$, an input $x$, features $f_1,\ldots,f_T$ and diameters $\bar{\delta}_1,\ldots,\bar{\delta}_T$,
if $\mathcal{A}$ is sound (but may be incomplete), then:
\begin{itemize}[nosep,nolistsep]
\item \tool is sound: at the end of the algorithm $I_{f_1,ds[1],\ldots,f_T,ds[T]}(x)$ is robust.
\item \tool is precise up to $\delta_\text{MIN}$
for each feature's maximal diameter.
\end{itemize}
\end{restatable}

\section{Evaluation}
\label{sec:eval}
In this section, we evaluate \tool. We begin with implementation aspects and optimizations and then present our experiments. 

\paragraph{Implementation}
We implemented \tool in Python\footnote{https://github.com/ananmkabaha/VeeP}.
It currently
supports neighborhoods defined by one or two features.
For the analyzer, it relies on GPUPoly~\cite{ref72}. 
It further builds on the idea of Semantify-NN~\cite{ref76} that encodes features as input layers with the goal of encoding pixel relations to reduce overapproximation errors. Semantify-NN encodes features using fully-connected and convolutional layers. For some features, this approach is infeasible for high-dimensional datasets because of the high memory overhead. To illustrate, denote the input dimension by $h\times w \times 3$. The HSL input layers, as defined in Semantify-NN,
map an (R,G,B) triple into a single value in the feature domain, resulting in a perturbed output of $h \times w$. This output is then translated back to the input domain. Namely, a fully-connected layer requires ${(h \times w) \times (h \times w \times 3)}$ weights. For ImageNet, where $h=w=224$, this layer becomes too large to fit into a standard memory (over 60GB). Instead, we observe that for some features the feature layer's weights are mostly zeros and thus this layer can be implemented using sparse layers~\cite{ref94,ref95}.
Our implementation sets $\delta_\text{MIN}=10^{-5}$ and $M=3$.
As optimization, it does not keep all previous examples, but only the required ones, which are dynamically determined. For example, for the neighborhood in \Cref{fig:fig_2d_multi}, \tool keeps only the examples at the top two rows.

\paragraph{Evaluation setup}
We trained models and ran the experiments on a dual AMD EPYC 7742 server with 1TB RAM and eight NVIDIA A100 GPUs. We evaluated \tool on four image datasets: MNIST~\cite{ref6} and Fashion-MNIST~\cite{ref103}, with images of size 28$\times$28, CIFAR-10~\cite{ref48}, with images of size 32$\times$32$\times$3, and ImageNet~\cite{ref99}, with images of size 224$\times$224$\times$3. %For the architectures,
We considered fully-connected, convolutional~\cite{ref104}, ResNet~\cite{ref5}, and AlexNet~\cite{ref97} models.
For MNIST and Fashion-MNIST, we used FC-5000x10, a fully-connected network with 50k neurons. For MNIST, we also used
a convolutional network SuperConv with 88k neurons (from ERAN's repository\footnote{https://github.com/eth-sri/eran}).
For CIFAR-10, we used ResNetTiny with 311k neurons (from ERAN) and ResNet18 with 558k neurons. For ImageNet, we used AlexNetTiny with 444k and AlexNet with 600k neurons. The last four models were trained with PGD~\cite{ref56}.
Since GPUPoly currently does not support MaxPool layers, we replaced them in AlexNet with convolutional ones (justified by~\cite{ref98}).
The CIFAR-10 models were taken from ERAN's repository, and we trained the other models.

\paragraph{Baseline approaches} We compare \tool to popular splitting approaches: branch-and-bound (BaB)~\cite{ref81,ref100,ref102,ref119,ref120,ref121,ref122} and uniform splitting~\cite{ref76,ref34,ref35}.
Any BaB technique starts by attempting to certify the robustness of the given neighborhood. If it fails, it splits the verification task into two parts and attempts to certify the robustness of each separately. If the certification fails again, BaB repeats the splitting process until all parts certify the original neighborhood.
The difference between BaB techniques is what neurons they can split and how they choose what to split. For example, some rely on heavy computations, such as solving a linear program~\cite{ref102,ref119}.
For our setting, where the split focuses on the input neurons and the input has low dimensionality, the \emph{long-edge} approach, which splits the input neuron with the largest interval, has been shown to be efficient~\cite{ref102}.
We thus compare to this approach.
 Uniform splitting splits a neighborhood into smaller neighborhoods of the same size, sufficiently small so the analyzer can certify them. Thus,
it requires a pre-determined split size (unlike \tool and BaB which adapt it during the execution).
For a fair comparison, we need to carefully determine this size:  
 providing a too small size will result in too long execution times (biasing our results), while providing a too large size will result in certification failures. Thus, we estimate the maximal
 split size which will enable the uniform splitting to certify successfully. To this end, before running the experiments, we run the following computation. For each neighborhood, we define several smaller neighborhoods. For each, we look for the maximal $\epsilon$ which can be verified by GPUPoly without splitting. Finally, we determine the split size of the uniform splitting to be the minimal value of $\epsilon$ across all these smaller neighborhoods.
 For a fair comparison, both baseline approaches were integrated in our system, i.e., they rely on GPUPoly and the feature layers described before.

 \paragraph{Experiments}
 We run two experiments: one limits the execution time with a timeout and measures the maximal certified diameter, and the other one measures execution time as a function of the certified diameter.
 In each experiment, we run multiple problem instances. In each instance, we provide each approach a network, an image, one or two features, and a target diameter (if there are two features, both have the same target diameter). We define the target diameter to be the diameter of the minimal feature adversarial example $\delta_{adv}$ (computed by a grid search).
 That is, we provide each approach an upper bound on the maximal certified diameter. We measure how close is the returned certified diameter to $\delta_{adv}$. Note that  our problem instances are challenging because the feature neighborhoods we consider are the largest possible.

\begin{table}[t]
\small
\begin{center}
\caption{\tool vs. branch-and-bound and uniform splitting over brightness, contrast, hue, saturation, and lightness neighborhoods, averaged over 50 images.}
\begin{tabular}{ l@{\hskip 0.075in}l@{\hskip 0.075in}l@{\hskip 0.075in}c@{\hskip 0.075in}c@{\hskip 0.075in}c@{\hskip 0.075in}c@{\hskip 0.075in}c@{\hskip 0.075in}c@{\hskip 0.075in}c }
    \toprule
Dataset  & Model      &                            &                  & \multicolumn{2}{c}{\tool} & \multicolumn{2}{c}{BaB}& \multicolumn{2}{c}{Uniform}\\
         &            &                            & $\delta_{adv}$   & $\delta_f$\% & $t[m]$ & $\delta_f$\%  & $t$[m] & $\delta_f$\%  & $t$[m]  \\
 \midrule
MNIST      & SuperConv  &   Brightness             &   0.61            &  100          &   0.5     &    100     &   1.16   &    98       &    4.1 \\
MNIST      & SuperConv   &   B\&C                  &  0.56             &  99          &  26.1      &    98         &  35.2         &      81        &   77.3    \\
MNIST      & FC 5000x10  &   Brightness            &  0.15             &  100          &  1.9     &  100        &    11.5    &  100         & 13.4   \\
MNIST      & FC 5000x10  &   B\&C                  &  0.134                &      94      &   54.5       &       59      &   86.4       &    62          &  81.8      \\
F-MNIST    & FC 5000x10  &   Brightness            &     0.3              & 100      & 3.5 & 100  & 15.5            &  100              &  27.9  \\
\midrule
CIFAR-10   & ResNetTiny   &   Brightness            &  0.42             &    100   &    7.9              &    100    &       32.1           &     89        &   60.6\\
CIFAR-10   & ResNetTiny   &   B\&C                  &  0.3             &    96      & 73.4        & 49    & 144.6            &     30         & 164.1   \\
CIFAR-10   & ResNetTiny   &   Hue                   &  3.36            &    99     & 27.5         & 62    & 59.1             &    77          &  48.94 \\
CIFAR-10   & ResNetTiny   &   Saturation            &  0.83            &    98     & 5.6          & 100   & 21.0             &  96     & 68.8   \\
CIFAR-10   & ResNetTiny   &   Lightness             &  0.39            &    100    & 10.8         & 100   & 45.9             & 76      & 32.6   \\
\midrule
ImageNet   & AlexNetTiny   &   Brightness            &      0.22       &  95        &   68.8               &   59     &     87.6       &      59            &  82.7  \\
ImageNet   & AlexNetTiny   &   Hue                   &  0.99   & 78        &  40.6                 &  25      &        67.4          &      37  & 68.1  \\
ImageNet   & AlexNetTiny   &   Saturation            &   0.39               &  97        &   27.7               &  79      &     69.0             &  71            & 74.9  \\
ImageNet   & AlexNetTiny   &   Lightness             &   0.16             &   93    &    64.8    &   17     &    83.4    &      52        & 71.4  \\
\bottomrule
\end{tabular}
    \label{tab:Semantify_NN}
    \end{center}
\end{table}

\paragraph{Maximal certified diameter given a timeout}
In the first experiment, we evaluate the maximal certified diameter of all approaches, given a timeout. The evaluated feature neighborhoods are defined by
 brightness (a linear feature) and contrast and HSL (non-linear features).
The contrast feature defines the brightness difference between light and dark areas of the image, and the HSL features are color space transformations, where hue defines the position in the color wheel, saturation controls the image's colorfulness and lightness the perceived brightness. We run \tool, BaB, and uniform splitting over the different models.
For most networks and neighborhoods, we let each splitting approach run on a single GPU for 1.5 hours.
For ResNet18, AlexNet, and the brightness and contrast (B\&C) neighborhoods of TinyResNet, we let each splitting approach run on eight GPUs for 3 hours.
We measure the execution time in minutes $t[m]$ and the maximal certifiable diameter $\delta_f$.
We compare $\delta_f$ to the diameter of the closest adversarial example in the feature domain $\delta_{adv}$ (for B\&C, we compare to $(\delta_{adv}$,$\delta_{adv})$).~\Cref{tab:Semantify_NN} reports our results for the smaller models. Each result is averaged on 50 images.
The results indicate that \tool proves on average at least 96\% of the maximal certifiable diameters in 29 minutes. The maximal diameters computed by the baselines are 74\%, for BaB, and 73\%, for uniform splitting. Their execution times are 54 minutes, for BaB, and 62 minutes, for uniform splitting.
\Cref{tab:results_big} reports our results for the two largest models, ResNet18 and AlexNet. Because of the long timeout, we focus on ten images and compare only to BaB.
 Our results show that \tool proves at least 96\% of the maximal diameters, while BaB proves 44\%. \tool's execution time is 98 minutes, whereas BaB is 160 minutes.

\begin{table}[t]
\small
\begin{center}
\caption{\tool vs. branch-and-bound over large models, averaged over 10 images.}
\begin{tabular}{ l@{\hskip 0.075in}l@{\hskip 0.075in}l@{\hskip 0.075in}c@{\hskip 0.075in}c@{\hskip 0.075in}c@{\hskip 0.075in}c@{\hskip 0.075in}c }
    \toprule
Dataset  & Model      &                            &                  & \multicolumn{2}{c}{\tool} & \multicolumn{2}{c}{BaB}\\
         &            &                            & $\delta_{adv}$   & $\delta_f$\% & $t[m]$ & $\delta_f$\%  & $t$[m]  \\
\midrule
CIFAR-10   & ResNet18     &   Brightness            &  0.41            &    100    & 88.4         & 58       &   150                \\
CIFAR-10   & ResNet18     &   Saturation            &  0.85            &    98     & 45.2         &  98      &   123              \\
\midrule
ImageNet   & AlexNet       &   Brightness            &   0.42             &  92    &    130     &      6       &     180         \\
ImageNet   & AlexNet       &   Saturation            &   0.56             &   100    &   67.3       &  52      &    165                \\
ImageNet   & AlexNet       &   Lightness             &   0.32             &   93    &  162       &   3     &      180             \\
\bottomrule
\end{tabular}
    \label{tab:results_big}
    \end{center}
\end{table}

\paragraph{Execution time as a function of the certified diameter}
In the second experiment,
we measure the execution time of every approach as a function of the certified diameter. In this experiment, there is no timeout and thus we focus on two models, ResNetTiny and AlexNetTiny, and two features: brightness and saturation.
For each network and a feature, we consider 50 images.
For each network, image, and a feature, the target diameter is the
 diameter of the closest adversarial example $\delta_{adv}$.
 We run all approaches until completion.
 During the execution of each approach, we record the intermediate progress, that is, the required time for certifying ${r}\cdot{\delta_{adv}}$ of the neighborhood, for ratio $r\in\{0.1,0.2,\ldots,0.8,0.9,0.95,0.98\}$.

 \begin{figure}[t]
    \centering
  \includegraphics[width=1.0\linewidth, trim=0 0 200 0, clip,page=6]{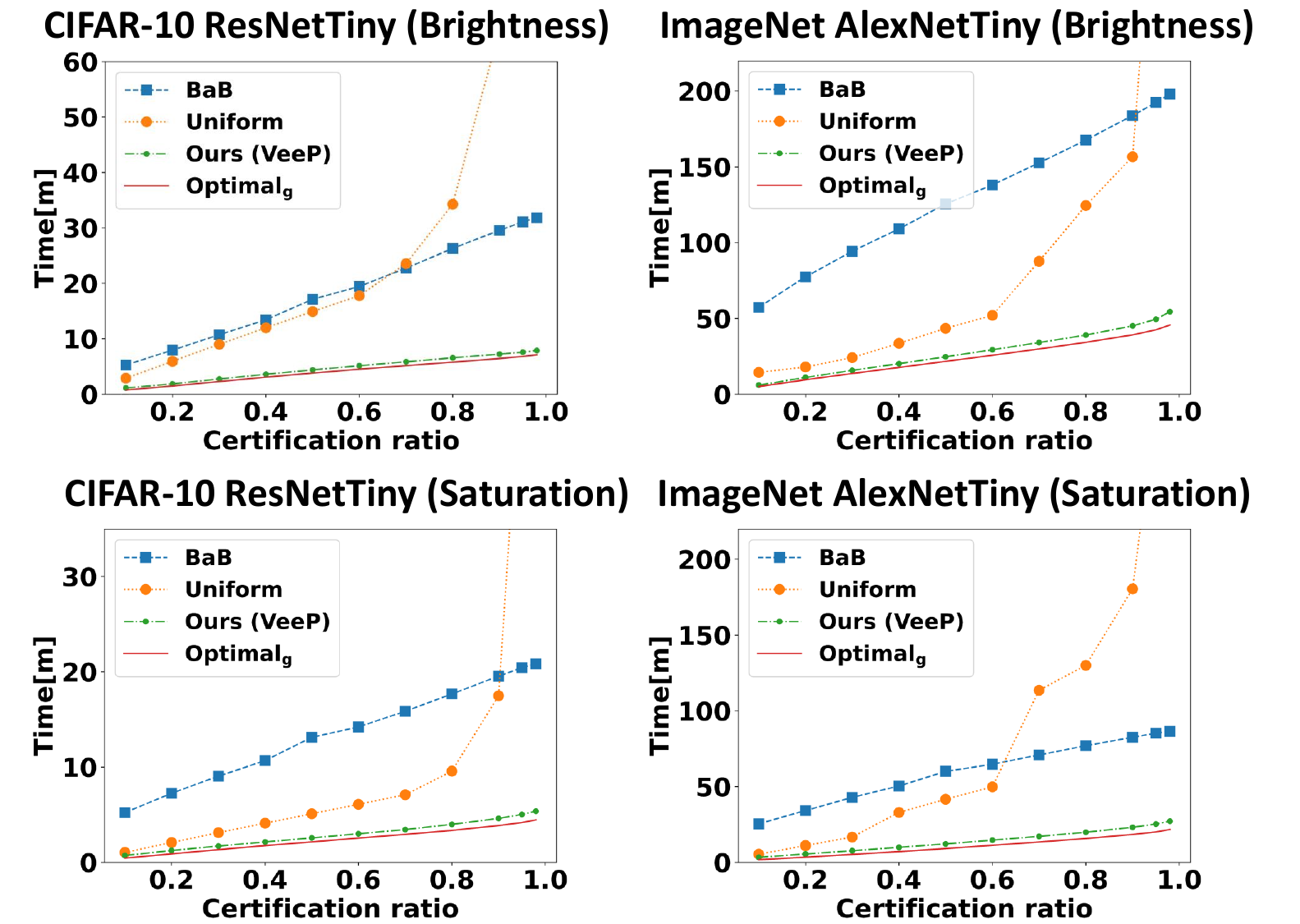}
    \caption{Comparison of \tool to uniform splitting, branch-and-bound, and a greedy optimal baseline, averaged over 50 images.}
    \label{fig:fig_scan}
\end{figure}

 \Cref{fig:fig_scan} shows the results of this experiment. It depicts the execution time in minutes of each approach as a function of $r$, i.e., the ratio of the certified diameter and the target diameter $\delta_{adv}$.
    Our results indicate that \tool provides acceleration of 4.4x compared to BaB and acceleration of 10.2x compared to uniform splitting. The figure demonstrates the main drawbacks of uniform splitting and branch-and-bound. On the one hand, choosing a large step size for uniform splitting can certify smaller ratios of the target diameter more quickly. On the other hand, for larger ratios, uniform splitting must use a smaller step size, which significantly increases the execution time. The results also show that BaB wastes a lot of time on attempts to certify too large neighborhood until converging to a certifiable split size. We note that both baseline approaches are sub-optimal since they do not attempt to compute the optimal split size. In contrast, \tool predicts the split sizes that minimize the execution time and thus performs better than the baselines.
   We validate \tool's optimality by comparing it to a theoretical greedy optimal baseline. The theoretical baseline ``knows'' (without any computation) the optimal step size for every verification step. To simulate it, before every verification step of the optimal baseline, we
   compute the optimal step size by running a grid search over the remaining diameter (i.e., $\bar{\delta}-\delta_x$). We then let the optimal baseline pick the diameter determined by the grid search.
    Note that this baseline is purely theoretical: we do not consider the execution time of running the grid searches as part of its execution time.
   Our results indicate that \tool's performance is very close to the theoretical baseline's performance, \tool is slower by only a factor of 1.2x.  The additional overhead of \tool stems from several factors: (1)~the time to estimate the predictors, (2)~the time to run the network on $f(x,\delta_x)$, and (3)~the inaccuracies of our predictors and correction steps.

\begin{figure}[t]
    \centering
  \includegraphics[width=0.95\linewidth, trim=0 310 460 0, clip,page=7]{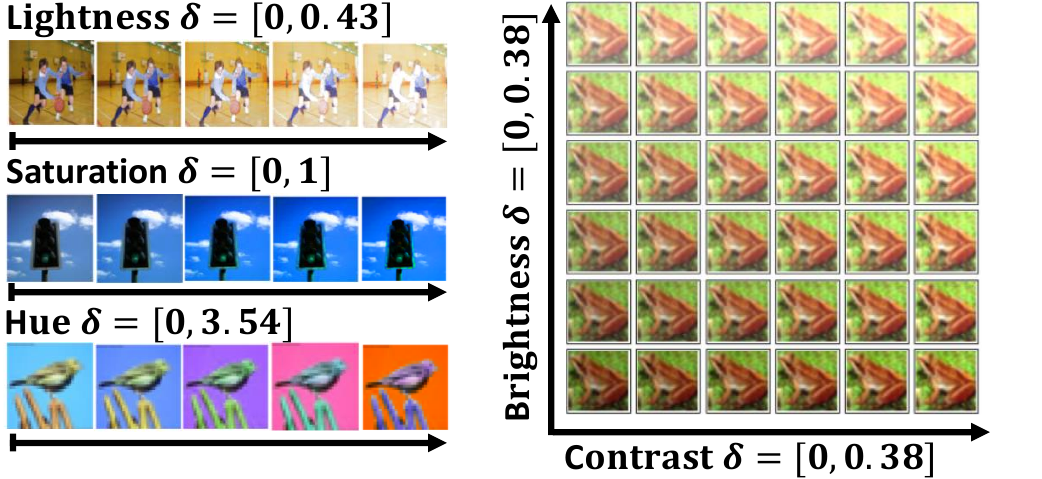}
    \caption{Examples of images in feature neighborhoods, certified by \tool.}
    \label{fig:fig_1d_2d}
\end{figure}

Lastly, we
exemplify how large the feature neighborhoods that \tool certifies are.
\Cref{fig:fig_1d_2d} shows four certified neighborhoods, defined by different features. 
For each, the figure shows the features, the range of the certified diameters, and several images generated by uniformly sampling from the certified range. The images are organized across the diameter axis, where the original image $x$ is at the origin. 
These examples demonstrate that the certified feature neighborhoods contain images that are visually different compared to the original image. Being able to certify large feature neighborhoods allows network designers understand the robustness level of their networks to feature perturbations.

\section{Related Work}
\label{sec:related_work}
In this section, we discuss the most closely related work to \tool. 
\paragraph{Network robustness and feature verification}
Many works introduce verifiers analyzing the robustness of $L_\infty$-balls, where each pixel is bounded by an interval~\cite{ref120,ElboherGK20,ref18,ref100,ref72,ref73,ref74,ref75,ref35,ref4,ref108}. Other works consider feature verification~\cite{ref76,ref34,ref35,ref41}. Earlier works on feature verification, focusing on rotations, brightness and contrast, translate feature neighborhoods into $L_\infty$ neighborhoods and then analyze them with existing verifiers~\cite{ref35,ref41}. Recent works encode the feature constraints into the verifier. One work relies on Monte Carlo sampling to overapproximate geometric feature constraints by convex linear bounds~\cite{ref34}. The bounds are refined by solving an optimization problem and then submitted to an existing verifier. Other work proposes an input layer that encodes the feature and is added to the original network~\cite{ref76}.
All works also employ uniform splitting. 

\paragraph{Splitting techniques}
To increase precision and scalability,
many verifiers rely on uniform splitting~\cite{ref76,ref34,ref35} or branch-and-bound (BaB)~\cite{ref81,ref100,ref102,ref119,ref120,ref121,ref122}. 
 Long-edge is a common BaB technique that splits the input with the largest interval~\cite{ref81,ref120}. Smart-Branching (BaBSB)~\cite{ref81} and Smart-ReLU (BaBSR)~\cite{ref102,ref100} rely on a fast computation to estimate the expected improvement of splitting an input or a neuron and then split the one maximizing the improvement. Filtered Smart Branching (FSB) extends BaBSR by bound propagation to estimate multiple candidates of BaBSR~\cite{ref119,ref100}.
  Another work relies on an indirect effect analysis to estimate the neuron splitting gain~\cite{ref122}. Others suggest to train GNNs via supervised learning to obtain a splitting strategy~\cite{ref121}. However, building the dataset and training the GNNs can be time consuming. 
    In contrast to BaB, which lazily splits inputs or neurons, \tool dynamically predicts the optimal split.

\paragraph{Feature attacks} Several adversarial attacks rely on semantic feature perturbations.
One work relies on HSV color transformations  (which is close to HSL)~\cite{ref77}. Other works link adversarial examples to PCA features~\cite{ref63,ref53,ref3}.
Other feature attacks include facial feature perturbations~\cite{ref9}, 
colorization and texture attacks~\cite{ref91}, features obtained using scale-invariant feature transform (SIFT)~\cite{ref10}, and semantic attribute perturbations using multi-attribute transformation models~\cite{ref90}.

\paragraph{Learning}
Our approach is related to several learning techniques. It is mainly related to active learning, where
a learner learns a concept by querying an oracle~\cite{ref116}. Active learning is suitable for tasks in which labeling a dataset is expensive~\cite{ref117}, for example real-life object detection~\cite{ref109}, crowd counting~\cite{ref111}, and image segmentation~\cite{ref110}. Similarly, in our setting,
querying the analyzer to obtain examples is expensive. 
Our setting is also related to online learning, where new data gradually becomes available. Online learning typically addresses tasks with time-dependent data~\cite{ref118}, e.g., visual tracking~\cite{ref112}, stock price prediction~\cite{ref113}, and recommendation systems~\cite{ref115}. 
In contrast, \tool's examples are not time-dependent. 
Our approach is also related to CEGIS and CEGAR. Counterexample-guided inductive synthesis (CEGIS) synthesizes a program by iteratively proposing candidate solutions to an oracle~\cite{ref124}. The oracle either confirms or returns a counterexample. 
Counterexample-guided abstraction-refinement (CEGAR)
is a program verification technique for dynamically computing abstractions capable of verifying a given property~\cite{ref126}. It begins from some abstraction to the program and iteratively refines it as long as there are spurious counterexamples. 
In contrast, \tool relies on recent examples, not necessarily counterexamples.

\section{Conclusion}
\label{sec:conclusions_and_discussion}
We presented \tool, a system for verifying the robustness of deep networks in neighborhoods defined by a set of features.
Given a neighborhood, \tool splits the verification process into a series of
verification steps, each aiming to verify a maximal part of the given neighborhood in a minimal execution time.
\tool defines the next verification step by constructing velocity and sensitivity predictors from previous steps and by considering recent failures. \tool is guaranteed to terminate and is sound and precise up to a parametric constant.
We evaluate \tool over challenging experiments: deep models for MNIST, Fashion-MNIST, CIFAR-10 and ImageNet,
and large feature neighborhoods, defined by the closest feature adversarial example.
Results show that the average diameter of the neighborhoods that \tool verifies is at least 96\% of the maximal certifiable diameter. Additionally, \tool provides a significant acceleration compared to existing splitting approaches: up to 10.2x compared to uniform splitting and 4.4x compared to branch-and-bound.
\subsubsection*{Acknowledgements.} We thank the reviewers for their feedback. This research was supported by the Israel Science Foundation (grant No. 2605/20). 
\bibliography{bib}
\appendix
\appendix
\newpage

\section{Examples using DeepPoly}\label{sec:appex}
\begin{figure}[t]
    \centering
  \includegraphics[width=1\linewidth, trim=0 260 0 0, clip,page=8]{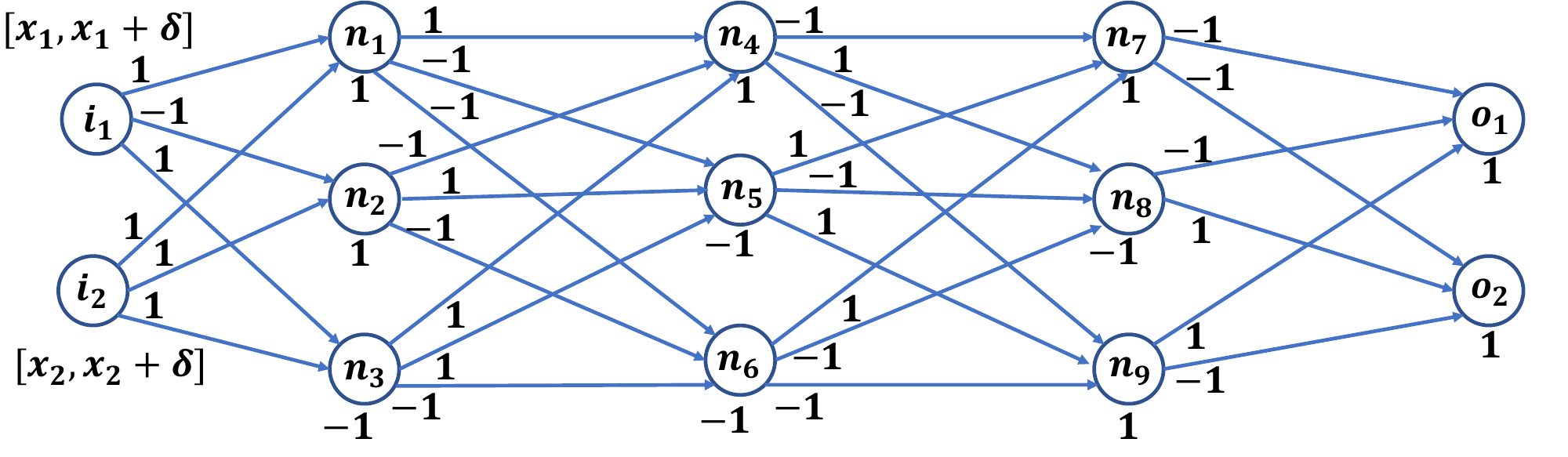}
    \caption{Small network example.}
    \label{fig::small_network_res2}
\end{figure}
In this section, we motivate our problem and Observation~\ref{obs2} with an example.
\paragraph{Motivation by example}
Consider
\Cref{fig::small_network_res2} showing an example of a small network comprising two inputs $[i_1,i_2]$, two outputs $[o_1,o_2]$, and three hidden layers. Each hidden layer contains three ReLU neurons, weights are depicted by the edges, and biases by the neurons.
For $x=[0.6,0.6]$, we aim at computing the maximal ${\delta}$ up to $\bar{\delta}=4$ whose $L_\infty$-ball is robust, i.e., $I_{{\delta}}(x)=\{(x_1+\delta_1,x_2+\delta_2)\mid 0\leq\delta_1,\delta_2\leq {\delta}\}$. For simplicity's sake, we do not consider a feature neighborhood, but rather a low dimensional $L_\infty$-ball.
Verifying this neighborhood with the DeepPoly~\cite{ref35} verifier fails due to loss of precision.
To succeed verifying, divide-and-conquer techniques split the neighborhood into smaller parts, which can be proven robust.
Assume we employ a vanilla branch-and-bound approach, using DeepPoly as the underlying verifier. The first call to the verifier is for the full neighborhood $I_{4}([0.6,0.6])$, which fails. It then splits the neighborhood into two neighborhoods: $I_{2}([0.6,0.6])$ and $I_{4}([2.6,2.6])$. Both neighborhoods are verified by DeepPoly, and thus it completes. The time waste is thus only for the first (original) neighborhood.
Assume we employ a vanilla uniform splitting approach, using DeepPoly as the underlying verifier, and that the predetermined number of splits is $m$. There is a time waste if $m=1$ or $m>2$. The larger the $m$, the more time is wasted. Instead, we aim to compute a series of diameters, $\delta_0=0,\delta_1,\ldots,\delta_j$ such that $\Sigma_{i} \delta_i =\bar{\delta}$, and such that (i)~each neighborhood $I_{f,\delta_{i+1}}(f(x,\delta_{i}))$ is successfully certified by the analyzer and (ii)~the execution time is minimal.

\paragraph{Observation~\ref{obs2} by example}
Next, we use the example of DeepPoly to provide an intuitive explanation to Observation~\ref{obs2}.
We begin with a short background.
DeepPoly bounds every neuron with lower and upper linear constraints of the form $n_{m,k}\geq L_{m,k} \cdot \hat{n}_{m,k}+B^L_{m,k}$ and $n_{m,k}\leq U_{m,k} \cdot \hat{n}_{m,k}+B^U_{m,k}$. Also, it computes real-valued lower and upper bounds $[l_{m,k}, u_{m,k}] \in \mathbb{R}^2$ bounding the neuron's possible values. An affine computation is captured precisely by the linear constraints $\hat{n}_{m,k}\leq b_{m,k}+\sum_{k'=1}^{k_{m-1}}{w}_{m,k,k'}\cdot{n}_{m-1,k'}$ and $\hat{n}_{m,k}\geq_{m,k}+\sum_{k'=1}^{k_{m-1}}{w}_{m,k,k'}\cdot{n}_{m-1,k'}$.
A ReLU neuron is captured precisely by the linear bounds if it is in \emph{stable state}. A ReLU neuron $n=ReLU(\hat{n})$, whose input is bounded by the interval $[\hat{l},\hat{u}]$, is in stable state if $\hat{u}\leq0$ or $\hat{l}\geq 0$. In stable state, the linear bounds of the ReLU neuron are either both equal to $n=0$ (if $\hat{u}\leq 0$) or both equal to $n=\hat{n}$ (if $\hat{l}\geq 0$).
A ReLU neuron which is not in stable state exhibits a non-linear computation, and thus the linear relaxation loses precision. Many linear relaxations have been proposed for ReLU, DeepPoly supports two possible relaxations (the choice among them is done independently for each non-stable ReLU neuron). As example, one of these relaxations is
$n\geq 0$ and
$n \leq \frac{\hat{u}}{\hat{u}-\hat{l}}\cdot (\hat{n}-\hat{l})$.

Now, we demonstrate Observation~\ref{obs2} on the example of
 DeepPoly. Recall:
\ftaf*

At high-level, this relation can be explained as follows. Consider the feature neighborhood of some step $s=(\delta_x,\delta)$, that is $I_{f,\delta}(f(x,\delta_x))$.
If $\delta=0$, i.e., the neighborhood contains a single input, then all ReLU neurons are stable and thus the linear approximation is exact.
 Assume we gradually increase $\delta$. As it increases, the number of unstable ReLU neurons increases. The relation between the number of unstable neurons and $\delta$ is a step function. We empirically observe that every increase in the number of unstable neurons leads to a linear decrease in the neighborhood sensitivity.
 That is, the neighborhood sensitivity is a piece-wise linear function. For large networks or neighborhoods, the number of unstable neurons quickly increases with the increase of $\delta$. Thus, each linear piece in the sensitivity function is very short. This leads to
 the observed exponential curve.
 \begin{figure}[t]
    \centering
  \includegraphics[width=1\linewidth, trim=0 275 0 0, clip,page=9]{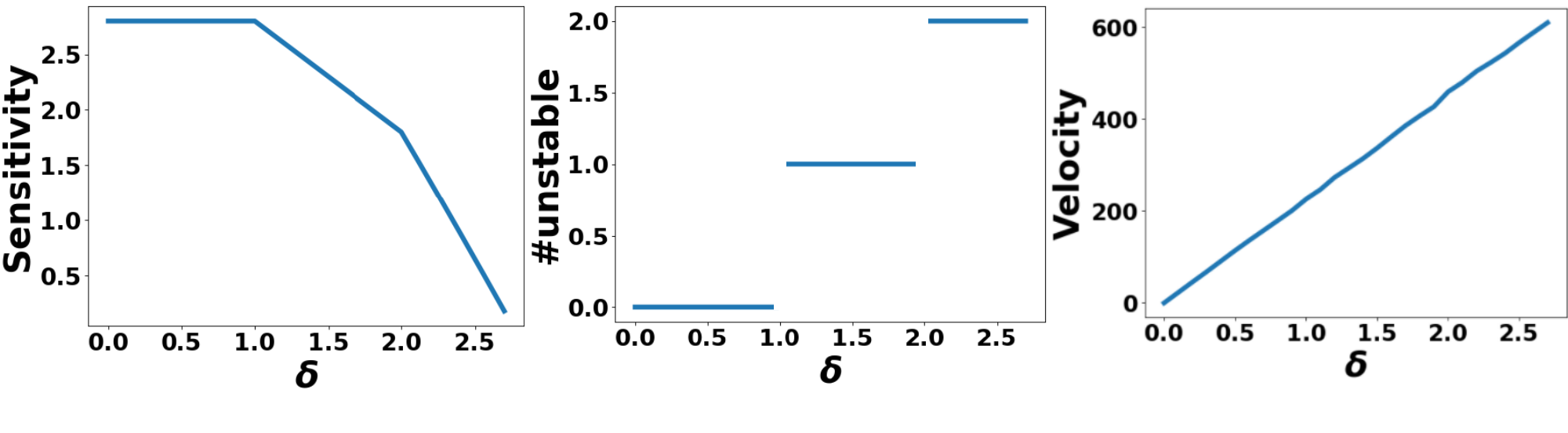}
    \caption{The proof sensitivity, number of unstable neurons, and proof velocity as a function of $\delta$ for the network in \Cref{fig::small_network_res2}.}
    \label{fig::small_network_res}
\end{figure}

 Next, we illustrate this relation on our example in \Cref{fig::small_network_res2}.
 Recall that in this example, we aim at computing the maximal ${\delta}\in [0,4]$ whose $L_\infty$-ball is robust, i.e., $I_{{\delta}}(x)=\{(x_1+\delta_1,x_2+\delta_2)\mid 0\leq\delta_1,\delta_2\leq {\delta}\}$.
 In this example, for every $\delta \in [0,1)$, all ReLU neurons are stable and the sensitivity is $S^D(I_{\delta}(x),2)=2.8$, when analyzed by DeepPoly.
    For $\delta \in[1,2)$, the neuron $n_2$ is the only unstable neuron. This is because $\hat{n}_2=1-i_1+i_2$ and thus $\hat{n}_2\in [1-\delta, 1+\delta]$. Consequently, DeepPoly sets the linear bounds of $n_2=ReLU(\hat{n}_2)$ to $n_2\geq 0$ and $n_2\leq \frac{1+\delta}{1-\delta}(\hat{n}_2+1-{\delta})$. The neighborhood sensitivity, for $\delta\in[1,2)$, is given by $S^D(I_{\delta}(x),2) = 3.8-\delta$. For $\delta\in[2,2.5]$, the neuron $n_5$ becomes unstable, too, because $\hat{n}_5\in[-\delta-2,\delta-2]$. The neighborhood sensitivity decreases to $S^D(I_{\delta}(x),2)  =6.2-2.2\cdot \delta$.
  \Cref{fig::small_network_res} visualizes the plots of the neighborhood sensitivity, the number of unstable neurons and the proof velocity as a function of $\delta$.
  We note that since this example shows a small network, the proof velocity is linear in $\delta$.

\section[Proofs]{Proofs}\label{sec:proofs}
In this section, we provide the proofs to our lemmas and theorems.

\ftb*
Proof.
We prove by showing
that the verification step sequence computed by \tool is finite.
Assume that for the given inputs, \tool generates a verification sequence $s_1,s_2,\ldots$. We show that its number of steps must be finite, and thus \tool is guaranteed to terminate, because $\mathcal{A}$ terminates and the rest of the computations are deterministic and finite.
Every verification step is associated with a result that $\mathcal{A}$ returns for the neighborhood defined by this step.
The number of steps in this sequence whose
results are \emph{robust} is finite and bounded by $\frac{\bar{\delta}}{\delta_\text{MIN}}$. This follows since the minimal diameter \tool submits to $\mathcal{A}$ is $\delta_\text{MIN}$. Next, we show
that every consecutive sequence of failed calls (\emph{non-robust} or \emph{unknown}) is finite. After we prove this, we can conclude that the overall verification sequence is finite.
Assume we have a consecutive sequence of failed calls.
Because of the validation test, every two adjacent diameters satisfy $\delta_{k+1}-c\leq\delta_{k}$, where $c>0$ is the constant that \tool decreases because of the failed call. The minimal diameter in this sequence is at least $\delta_\text{MIN}$. Thus, this sequence cannot be infinite, because if one of the diameters in this sequence is $\delta_\text{MIN}$, and $\mathcal{A}$ fails for this diameter, then the execution of \tool terminates.

\fta*
Proof. We show it
by induction on the length of the verification sequence $k$.

Base, $k=1$: In this case, \tool submits to $\mathcal{A}$ a single neighborhood $I_{f,\bar{\delta}}(x)$ and returns $\delta_x=\bar{\delta}$ or $\delta_x=0$, based on its response.
If $\mathcal{A}$ returns \emph{robust}, then $\delta_x=\bar{\delta}$ and
by the soundness of $\mathcal{A}$, it follows that \tool is sound.
If $\mathcal{A}$ does not return \emph{robust}, then because there is only a single step, it must be that $\bar{\delta}=\delta_\text{MIN}$. In this case,
the neighborhood is small enough such that we assume that $\mathcal{A}$ does not lose precision, and therefore we assume that there is an adversarial example for $x'=f(x,\hat{\delta})$, where $\hat{\delta}\in (0,\delta_\text{MIN}]$.

Step: Assume the sequence length is $k$. Then, using the first $k-1$ verification steps, \tool verified the neighborhood $I_{f,\delta'_{x}}(x)$.
By the induction hypothesis, $I_{f,\delta_{x}'}(x)$ is robust. In the last step, \tool submits to $\mathcal{A}$ the neighborhood $I_{f,\delta_k}(f(x,\delta_{x}'))$, where $\delta_k\leq \bar{\delta}-\delta_{x}'$ and returns $\delta_x=\delta_{x}'+\delta_k$ or $\delta_x=\delta_{x}'$, based on
 its response. If $\mathcal{A}$ returns \emph{robust}, then by the soundness of $\mathcal{A}$ and since $I_{f,\delta_{x}'+\delta_k}(x)=I_{f,\delta_k}(f(x,\delta_{x}'))\cup I_{f,\delta_{x}'}(x)$, it follows that $I_{f,\delta_{x}'+\delta_k}(x)$ is robust.
Since this is the last verification step of \tool,
it follows that $\delta_{x}'+\delta_k=\bar{\delta}$.
If $\mathcal{A}$ does not return \emph{robust}, then $\delta_k=\delta_\text{MIN}$. In this case,
the neighborhood is small enough such that  we assume that $\mathcal{A}$ does not lose precision,
 and therefore we assume that there is an adversarial example for $x'=f(x,\hat{\delta})$, where $\hat{\delta}\in (\delta_x',\delta_x'+\delta_\text{MIN}]$.

\ftc*
 Proof sketch. The proof shows that the number of distinct iterations is finite and that the termination condition must be satisfied at one of these iterations (i.e.,
there are no identical iterations).
The proof that the number of iterations is finite is similar to~\Cref{lem2}. We show that the maximal number of verified steps is $(\frac{\bar{\delta}}{\delta_\text{MIN}})^T$ and that any sequence of consecutive failed steps is finite. This follow from a simple extension of the reasoning in the proof of \Cref{lem2}, i.e., we show:
(1)~\texttt{compute\_next\_offsets} returns an offset for which \tool defines an (unproven) neighborhood whose hyper-volume is at least
 $(\delta_\text{MIN})^T$ and (2)~upon failure for diameter $\delta=\delta_\text{MIN}$, \texttt{compute\_next\_offsets} moves to another feature direction; thus, after failing $T$ times with diameter $\delta_\text{MIN}$,
 the sequence terminates. Thus, the number of distinct iterations is finite.
 To show that the termination condition is met at one of these iterations, we split to cases.
  If \tool certifies the target diameter of every feature, then we show that \texttt{ds} contains
 all target diameters, and thus the loop terminates. This follows because if all target diameters
 are certified,
 \texttt{compute\_certified\_neighborhood\_vertices} returns the vertices bounding the maximal hyper-rectangle,
 and thus $\forall i. ds\_curr[i]=\bar{\delta}_i$. In the first iteration it happens,
 the norm of \texttt{ds\_curr} must be
greater than the current \texttt{ds}, and thus \texttt{ds} is set to \texttt{ds\_curr} and the loop terminates.
Otherwise, there is a sequence of $T$ failed iterations with diameter $\delta_\text{MIN}$,
thus the termination condition is satisfied.

\ftd*
Proof is similar to the proof of \Cref{thm}, but consists of $T$ composed inductions.

\section[Offsets]{Computing Offsets}\label{sec:appex_offsets}
\reusefigure[t]{fig:fig_2d_multi}

In this section, we describe how \tool computes the next offsets.
\tool computes the maximal diameters feature-by-feature.
Assume that \tool is currently computing the maximal diameter of $f_i$, after computing the maximal diameters of $f_1,\ldots,f_{i-1}$: $\delta_1',\ldots,\delta_{i-1}'$.
To compute the maximal diameter of the $i^\text{th}$ feature, \tool
 computes the maximal robust $i$-dimensional neighborhood of the first $i$ features.
 As in \Cref{sec:onefeat}, \tool computes the maximal diameter $\delta_i'$ step-by-step.
Unlike \Cref{sec:onefeat}, not every step directly increases $\delta_i'$. This is because
 increasing $\delta_i'$ by $\delta$ requires
verifying a hyper-rectangle of dimensions $\delta_1'\times \ldots \times \delta_{i-1}'\times \delta$, which may be too large for the analyzer $\mathcal{A}$ to succeed. Instead, \tool
covers the maximal $i$-dimensional neighborhood by a series of hyper-cubes, whose diameter is predicted (as described in~\Cref{sec:onefeat}).
The series follows a feature-by-feature order (from $1$ to $i$), thereby
guaranteeing that all inputs in the maximal $i$-dimensional neighborhood have been covered by some hyper-cube whose analysis succeeded.

The hyper-cubes' order is as follows (in the following we use verification steps and hyper-cubes interchangeably).
When starting the computation of $\delta_i'$, \tool computes the initial offsets based on the already certified neighborhoods (except for $i=1$, whose initial offsets are all zeros). This is obtained by finding the earliest hyper-cube forming a vertex on the $i$-dimensional boundary of the certified region, such that the vertex's $i^\text{th}$ offset is within $(0,\bar{\delta}_i)$. It sets the initial offsets to that vertex's offsets. Then, \tool constructs hyper-cubes advancing in the direction of the first feature (top row of \Cref{fig:fig_2d_multi}(a), for the brightness feature, and top row of \Cref{fig:fig_2d_multi}(b), for the contrast feature).
 When reaching the maximal diameter (i.e., a hyper-cube whose diameter is $\delta_\text{MIN}$ fails or a hyper-cube succeeds and reaches $\bar{\delta}_1$), \tool looks for the earliest hyper-cube
forming a vertex on the $2$-dimensional boundary of the certified region, such that the vertex's second offset is within $(0,\bar{\delta}_2)$
 (for the contrast feature, this is the leftmost cube, top row, \Cref{fig:fig_2d_multi}(b)). It sets
 the next hyper-cube's offsets to that vertex's offset.
 Then, \tool again constructs hyper-cubes advancing the first feature's offset,
  until (again) reaching its  maximal certified diameter (rightmost square, top row, \Cref{fig:fig_2d_multi}(c)).
   Every time the first feature's reaches a maximal diameter, \tool (again) looks for the earliest hyper-cube forming a vertex on the $2$-dimensional boundary of the certified region,
  such that the vertex's second offset is within $(0,\bar{\delta}_2)$. Then, it continues from that vertex's offsets and constructs hyper-cubes advancing the first feature's offset.
  If, during the analysis, a hyper-cube fails for $\delta>\delta_\text{MIN}$, the next hyper-cube begins from the same offsets.
  At some point of this computation, \tool reaches the
  maximal diameter of the \emph{second} feature. This is either when
    (1)~all vertices' second offset is at least $\bar{\delta}_2$ or (2)~$\mathcal{A}$ fails for a hyper-cube whose offsets equal the offsets of a vertex and the hyper-cube's diameter is $\delta_\text{MIN}$. Then, \tool repeats a similar process for the third feature: it looks for the earliest hyper-cube
forming a vertex on the $3$-dimensional boundary of the certified region, such that the vertex's third offset is within $(0,\bar{\delta}_3)$. Then, it constructs hyper-cubes as before: the hyper-cubes advance the first offset, when a hyper-cube reaches the first feature's maximal diameter, the second offset advances, and so on until a hyper-cube reaches the second feature's maximal diameter.
Then, \tool advances the third offset (by looking for the corresponding vertex) and repeats this process until reaching the third feature's maximal diameter.
  Generally, when \tool reaches the maximal diameter of the $({j-1})^\text{th}$ feature, for $j\leq i$, \tool repeats the same process: it looks for the earliest hyper-cube
forming a vertex on the $j$-dimensional boundary of the certified region, such that the vertex's $j^\text{th}$ offset is within $(0,\bar{\delta}_j)$ and constructs hyper-cubes advancing the previous features by the order described. That is, whenever reaching the $(j-1)^\text{th}$ feature's maximal diameter, \tool advances the offset of the $j^\text{th}$ feature.

Based on this order, the procedure \texttt{compute\_next\_offsets} returns the offsets of the next hyper-cube.
It computes the offsets by identifying which feature's diameter should be advanced. This is computed from the set of examples defining the vertices of the certified region. Examples that do not form a vertex are ignored.
Given this set of examples, \tool identifies the feature to advance by locating the minimal feature (with respect to its index in $f_1,\ldots,f_T$) whose offset is smaller than the corresponding target diameter. For example, after the two steps at the top row of \Cref{fig:fig_2d_multi}(c), \texttt{compute\_next\_offsets} identifies that the feature to advance is $f_1$ since it is the minimal feature whose offset has not reached the target diameter.
Note that the examples at the two bottom rows are ignored since neither forms a vertex on the certified region.
After identifying the feature to advance $i_f$, \texttt{compute\_next\_offsets} looks for the earliest hyper-cube (i.e., example)
forming a vertex on the $i_f$-dimensional boundary of the certified region, such that the vertex's $i_f^\text{th}$ offset is within $(0,\bar{\delta}_{i_f})$.
It sets the new offset to that vertex's offset.
A couple of notes on this computation.
First, this computation is correct even if \tool will not reach the target diameter of the chosen feature.
Second, the \texttt{predict} function (Line 8, \Cref{alg:main}) may slightly refine the offsets returned by \texttt{compute\_next\_offsets} based on the predicted $\delta$. The refinement is computed from the vertices of the certified region $BV_1,...,BV_T$ and the predicted diameter $\delta$ as follows.
For every $f_i$ between the feature $f_{i_f}$ (which \texttt{compute\_next\_offsets} identified) and $f_T$, it computes $v = \min\{BV_{i_f}[i] \mid BV_{i_f}[i] \in (\text{offsets}[i_f],\text{offsets}[i_f]+\delta]\}$ (if the set is empty, then $v=\bot$). Accordingly, it updates the offsets: $\text{offsets}[i_f] = (v\:\neq\:\bot)? \: v[i_f]: \: \text{offsets}[i_f]$. For example, the offsets returned by \texttt{compute\_next\_offsets} for the third hyper-cube at the top row of \Cref{fig:fig_2d_multi}(b) are
$(0.035,0.018)$. Then, \texttt{predict} predicts $\delta=0.019$ and afterwards checks the vertices $(0.053,0.018)$ (determined by the third hyper-cube, bottom row) and $(0.053,0.017)$ (determined by the fourth hyper-cube, bottom row). To ensure that no input is missed, \texttt{predict} refines the second offset to $0.017$.
That is, the final offsets of this step are $(0.035, 0.017)$.

\Cref{alg:compute_offsets1} summarizes \texttt{compute\_next\_offsets}'s operation.
It takes as input all examples \texttt{Ex} (each corresponds to a hyper-cube) and the target diameters. \texttt{compute\_next\_offsets} relies on three static variables.
The first is an array of size $T$, called $i_e$. For every index $j$, the value at $i_e[j]$ equals to the minimal index of \texttt{Ex} containing a vertex on the $j$-dimensional boundary of the certified region. In other words, for feature $j$, every example in \texttt{Ex} between index $1$ and $i_e[j]-1$ can be ignored (as explained before).
Initially, all indices are $1$ (no example is ignored).
The second static variable is a boolean flag \texttt{finish}, indicating whether all maximal diameters have been reached; initially, it is \texttt{False}. The third static variable $ex_{\delta_\text{MIN}}$ maintains the first example that failed for $\delta_\text{MIN}$ (initially, it is $\bot$).
The goal of \texttt{finish} and $ex_{\delta_\text{MIN}}$ is to provide an optimization for the following special case. It may happen
that \texttt{compute\_next\_offsets} identifies it has reached all maximal diameters \emph{without} failing $T$ times or reaching all target diameters. It may happen if as a byproduct of the analysis of preceding features, succeeding features have reached their maximal diameters. In this case, \texttt{compute\_next\_offsets} repeatedly returns a hyper-cube that will fail for any $\delta \geq \delta_\text{MIN}$, until \tool fails for $T$ consecutive times.
When \texttt{compute\_next\_offsets} detects this case, it sets \texttt{finish=True} and from here on the offsets of $ex_{\delta_\text{MIN}}$ are returned until \tool completes (Line 1). We note that if $ex_{\delta_\text{MIN}}=\bot$, it returns the last example's offsets.
 Besides this special case's optimization, the operation of \texttt{compute\_next\_offsets} is as described.
 It begins by identifying the feature $i_f$ whose diameter should be advanced
and the example \texttt{ex} whose vertex's offsets are used to define the next offsets.
It initially sets $i_f=1$.
If the last step failed (i.e., its sensitivity is not positive)
and the last diameter is greater than $\delta_{\text{MIN}}$, then
the returned offsets are those of the last step (Line 4). In this case, the next iteration of \tool will predict a smaller diameter.
If the last step failed and the last diameter is $\delta_{\text{MIN}}$, then
\tool has reached the maximal diameter of the current feature.
This feature is identified by finding the minimal feature whose offset is not zero (Line 5).
In this case, $i_f$ increases by 1 to continue to the next feature (Line 6).
Then, if $ex_{\delta_\text{MIN}}$ is $\bot$, it is set to this failed example (Line 7).
After the if-block, \texttt{compute\_next\_offsets} looks for the minimal unfinished feature $i_f$ within the (current) $i_f$ and $T$. This feature
is the minimal one whose earliest hyper-cube forming a vertex (on the $i_f$-dimensional boundary of the certified region) satisfies that the vertex's $i_f^\text{th}$ offset is within $(0,\bar{\delta}_{i_f})$ (Lines 8-10).
If there is an unfinished feature $i_f$ (Line 11), \texttt{compute\_next\_offsets} determines the new offsets based on the corresponding vertex. To this end, it finds the vertex's example \texttt{ex} (Lines 12-13);
if there is no such example, it takes the last example. Then,
\texttt{compute\_next\_offsets} defines the returned offsets to those of that vertex.
That is,
the new offsets are those of \texttt{ex}, except that, if \texttt{ex} succeeded, $i_f$'s offset increases by \texttt{ex}'s diameter $\delta$ (Lines 14-15). Lastly,
\texttt{compute\_next\_offsets} removes examples that no longer form vertices on any $j$-dimensional boundary of the certified region, for $j\leq i_f$ (Line 16).
If all features have finished (Line 17), \texttt{compute\_next\_offsets} sets \texttt{finish} to \texttt{True} (Line 18) and returns the offsets of $ex_{\delta_\text{MIN}}$, if exists, or those of the last example otherwise (Line 19).

\begin{algorithm}[t]
\Parameter{$i_e$ = $[1,\ldots,1]_T$, finish = False, $ex_{\delta_{MIN}}$ = $\bot$}
\DontPrintSemicolon
   \lIf{finish}{\Return{$ex_{\delta_\text{MIN}} \neq \bot ? \:  ex_{\delta_\text{MIN}}.\text{offsets} : Ex[-1].\text{offsets}  $}}
   $i_f = 1$\;
  \caption{compute\_next\_offsets($Ex$, $\bar{\delta}_1,\ldots,\bar{\delta}_T$)}\label{alg:compute_offsets1}
    \If { $Ex[-1].S\leq0$}{
        \lIf { $Ex[-1].\delta>\delta_\text{MIN}$}{\Return{$\text{Ex[-1].offsets}$}}
         $i_f$ = $\text{min}\{i\in \{1,\ldots,T\} \mid Ex[-1].\text{offsets}[i] \neq 0\}$ \tcp*{Current feature}
         $i_f$ = $\min\{i_f+1,T\}$ \tcp*{Continue to next feature}
         $ex_{\delta_\text{MIN}}$= ($ex_{\delta_\text{MIN}}$=$\bot$? $Ex[-1]$ : $ex_{\delta_\text{MIN}}$)
  }

    \For(\tcp*[h]{Find next unfinished feature}){ ; $i_f \leq T$; $i_f$++}{
      \If{$\exists e'\in Ex[i_e[i_f]:-1]:\ ( e'.\text{offsets}[i_f]+e'.\delta< \bar{\delta}_{i_f})\land (e'.S>0)$}{
        break
      }

    }

  \eIf(\tcp*[h]{ Found a feature that did not finish}){$i_f \neq T+1$}{
  $Ex_{vertex}$ = $\{e'\in Ex[i_e[i_f]:-1] \mid (e'.\text{offsets}[i_f]+e'.\delta< \bar{\delta}_{i_f}) \land (e'.S>0) \}$\;
  $ex = Ex_{vertex}\:\neq\:\emptyset ? \: Ex_{vertex}[1] \: : \: Ex[-1]$\;
  $\text{offsets} = ex.\text{offsets}$\;
  $\text{offsets}[i_f] = ex.S>0?\:\text{offsets}[i_f] + ex.\delta\: : \text{offsets}[i_f]  $\tcp*{Vertex's offsets}
  $i_e[1:i_f]=Index(Ex[-1])$ \tcp*{Remove examples not forming vertices}
  }(\tcp*[h]{Completed all features}){

   finish = True\;
   $\text{offsets} = ex_{\delta_\text{MIN}} \neq \bot ?\:  ex_{\delta_\text{MIN}}.\text{offsets} : Ex[-1].\text{offsets}  $\;

  }
\Return{$\text{offsets}$}
\end{algorithm}

\end{document}